\documentclass{article}

\usepackage[preprint]{neurips_2026}

\newcommand{\bench}{\textbf{WorldBench}}

\usepackage[utf8]{inputenc} 
\usepackage[T1]{fontenc}    
\usepackage{hyperref}       
\usepackage{url}            
\usepackage{booktabs}       
\usepackage{amsfonts}       
\usepackage{nicefrac}       
\usepackage{microtype}      
\usepackage{xcolor}         
\usepackage{graphicx}
\usepackage{cleveref}
\usepackage{caption}
\usepackage{multirow}
\usepackage{gensymb}
\usepackage{adjustbox} 
\usepackage[table,xcdraw]{xcolor} 
\usepackage{pifont}
\title{\bench : Benchmarking Physical Understanding of World Models by Isolating Physics Concepts}

\author{%
  Rishi Upadhyay \\
  UCLA \\
  \And 
  Howard Zhang \\
  UCLA \\
  \And 
  Jim Solomon \\
  UCLA \\
  \And 
  Ayush Agrawal \\
  UCLA \\
  \And 
  Yunhao Ba \\
  Sony AI \\
  \And 
  Alex Wong \\
  Yale University \\
  \And 
  Celso M de Melo \\
  DEVCOM Army Research Laboratory 
  \And 
  Achuta Kadambi \\
  UCLA \\
}

\begin{document}

\maketitle


\begin{abstract}
Recent advances in generative foundational models, often termed "world models," have propelled interest in applying them to critical tasks like robotic planning and autonomous system training. For reliable deployment, these models must exhibit high physical fidelity, accurately simulating real-world dynamics. Existing physics-based video benchmarks, however, suffer from entanglement, where a single test simultaneously evaluates multiple physical laws and concepts, fundamentally limiting their diagnostic capability. We introduce WorldBench, a novel video-based benchmark specifically designed for concept-specific, disentangled evaluation, allowing us to rigorously isolate and assess understanding of a single physical concept or law at a time. To make WorldBench comprehensive, we design benchmarks at two different levels: 1) an evaluation of intuitive physical understanding with higher level concepts such as object permanence or scale/perspective, and 2) an evaluation of low-level physical constants and material properties such as friction coefficients or fluid viscosity, allowing to measure excatly how far from reality generated videos are. When SOTA video-based world models are evaluated on WorldBench, we find specific patterns of failure in particular physics concepts, with all tested models lacking the physical consistency required to generate reliable real-world interactions. Through its concept-specific evaluation, WorldBench offers a more nuanced and scalable framework for rigorously evaluating the physical reasoning capabilities of video generation and world models, paving the way for more robust and generalizable world-model-driven learning.
\end{abstract}    

\section{Introduction}

As interests in robotics and embodied AI tasks grows, there has been a growing focus on leveraging video-generation based and similar world models as sources of synthetic data to replace physics simulators which can be slow, limited, and lacking in diversity. In contrast, generation based synthetic data offers the promise of diverse, high-quality data generated on the fly as needs arise. For example, there has been specific interest in using video models to generate long-tailed data, capturing cases that are not frequently seen in real life.

However, for the synthetic data generated by these models to be truly useful, it must closely match real world data in terms of both visual quality and the dynamics and motion of objects. Recent world foundation models such as NVIDIA's Cosmos~\cite{agarwal2025cosmos} promise to model both correctly, but prior work has shown this is not always the case~\cite{li2025pisa, bansal2024videophy}. For these models to truly be useful as synthetic data generators, they must very closely match real world dynamics, down to the exact motion of objects and the physical constants that govern that motion. Therefore, rigorously evaluting these models ability as data generators requires benchmarks that are designed and focused on probing physical understanding at a concept-specific and objective level.

In this paper, we introduce \bench, a novel benchmark designed to rigorously evaluate the physical reasoning capabilities of WFMs through video prediction. Our testing framework uses fine-grained categories of evaluation to fill in a critical gap in in the current research landscape. The video-based output requires models to accurately forecast the physically plausible evolution of full visual scenes over time, providing a more robust signal than previous binary metrics. To achieve this nuanced assessment while ensuring repeatable, interpretable outcomes, we design simplified, yet physically rich and visually realistic scenes.

Our benchmark is split into two subsets, both of which address crucial aspects of physics understanding. The intuitive physical understanding subset targets four critical principles: motion physics, object permanence, support relations, and scale/perspective. This subset is meant to benchmark a model's ability to generate plausible dynamics governed mostly by their respective core principles (e.g. ball rolling behind pillars, object moving towards the camera, etc.). 

\bench~also includes a physical parameter estimation subset. This feature requires the video output to accurately adhere to specific, known physical parameters that govern the scene's dynamics, such as gravitational acceleration, fluid viscosity, and friction coefficients. By enforcing adherence to these measurable parameters, we provide a definitive, quantifiable metric for assessing a model's true grasp of each underlying physical law. This is also crucial for data generation: a physics simulator which generates incorrect gravity cannot be used to generate data for a model that needs to operate in the real world.

Our benchmark allows for a significantly more nuanced and detailed assessment of physical laws than previous work, including a more fine-grained diagnosis of which concept a model underperforms on, and an accurate representation of object dynamics (velocity, acceleration, rotation, deformation, and occlusion). The first subset's focus on specific, real-world physics properties, such as object permanence and scale/perspective—which are absent in prior benchmarks like Physion—ensures that \bench~provides a richer signal of how close these models are to truly learning and understanding real-world dynamics. The second subset's focus on real-world physical parameters also enables us to directly measure the physical accuracy of world models, which is particularly critical if these models aim to be synthetic data generators.

Using \bench, we extensively test state-of-the-art world models, including the Cosmos architecture, revealing substantial gaps in physical consistency and generalization when compared to both real-world captured and simulated, physically accurate expectations. Importantly, we find that models tend to generate visually realistic scene evolutions (e.g. ball follows parabolic trajectory), but fail to adhere to physical parameters (e.g. ball accelerates downward at $9.8 \frac{m}{s}^2$). By leveraging video outputs rather than binary selection tasks, our benchmark provides a way to quantitatively disambiguate between visually plausible and physically accurate model outputs. Our results highlight the critical limitations of current world model architectures and motivate further research into physically grounded learning. 

\noindent
Below is a summary of our key contributions:
\begin{itemize}
    \item We introduce a video-based benchmark carefully designed to ensure each sample evaluates a single physics concept or constant, allowing for more fine-grained understanding of world model physics understanding.
    \item We provide a dataset and codebase to estimate exact physical constants such as $g$ from video models, providing an absolute, meaningful metric to judge model performance as opposed to subjective (e.g. based on VLM judgement) or relative metrics.
    \item We perform an empirical analysis of SoTA WFMs and Image-to-Video models to identify concept-specific shortcomings and gaps in physical understanding, providing directions for future improvement.
\end{itemize}

\begin{figure*}
    \centering
    \includegraphics[width=.7\textwidth]{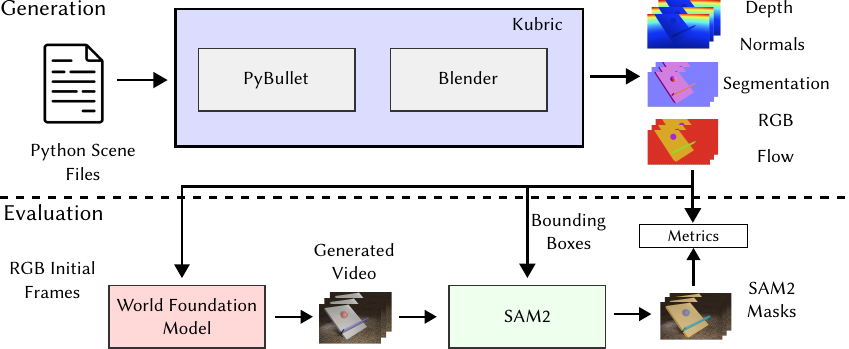}
    \caption{\textbf{Overview of our generation and evaluation process}. For generation (top), we use Kubric, which uses PyBullet and Blender under the hood. During evaluation (bottom), we first pass the initial frames of the generated video to the world foundation model which completes the video. The completed video is passed to SAM2 along with bounding boxes based on ground truth masks. The segmentations outputted by SAM2 are compared to ground truth segmentations to obtain the final metrics.}
    \label{fig:overview}
\end{figure*}

\section{Related Work}
\subsection{World Foundation Models}
A significant body of work has emerged around "world models", models that can understand and predict the real world, in recent years. Initial work in this space focused on vision-language models~\cite{liu2024world, micheli2022transformers, robine2023transformer}, but recent work has been on using video generation models~\cite{alonso2024diffusion}. These models typically leverage transformer architectures and either latent diffusion models~\cite{blattmann2023stable,blattmann2023align,esser2023structure,ho2022video,hong2022cogvideo,ceylan2023pix2video,luo2023videofusion,singer2022make,wang2023videofactory} or auto-regressive models~\cite{deng2024autoregressive, weng2024art, xie2024progressive, yin2024slow, wu2022nuwa,weissenborn2019scaling,yan2021videogpt,wu2021godiva,rakhimov2020latent} to achieve temporally consistent video synthesis. However, while these models are able to generate visually realistic and aesthetically pleasing outputs, there has also been a recent growth of research surrounding physically accurate generations. The recent Cosmos~\cite{agarwal2025cosmos, ali2025world} aims to be a ``world foundation model", which can output temporally and physically accurate videos that can be leveraged for training downstream AI models that interact with the physical environment. Cosmos can generate these videos using either a transformer-based autoregressive model or a transformer-based diffusion model, training a large corpus of over 100M video clips, labeled by numerous different vision-language models~\cite{vaswani2017attention}. Similarly, other models such as Genie~\cite{bruce2024genie}, also attempt at creating a ``world foundation model" capable of generating physically accurate interactive environments. These "world foundation models" claim to be physically accurate enough for their outputs to be used as simulated data, but little to no evaluations have been developed so far to validate this claim.

\subsection{Image-to-Video Models}
A closely related and highly active area of research is Image-to-Video (I2V) generation, which focuses on synthesizing a video sequence from a single static image input. The fundamental challenge in I2V models is temporal consistency: ensuring that the generated frames maintain the identity, structure, and appearance of the initial image while introducing plausible motion and scene evolution.

Modern I2V models typically ``inflate" standard image diffusion models with a temporal dimension, using temporal convolution or attention layers~\cite{blattmann2023stable,blattmann2023align,gu2023reuse,guo2023animatediff,he2022latent,wang2023modelscope,hong2022cogvideo,yang2024cogvideox,wan2025wan,hunyuanworld2025tencent}. Notable models include CogVideoX~\cite{hong2022cogvideo,yang2024cogvideox}, which leverage a 3D-VAE for enhanced compression and an expert transformer for better text-video alignment, as well as WAN~\cite{wan2025wan} which leveraged large-scale data training and a spatio-temporal VAE to achieve SOTA results on previous benchmarks.

While I2V models excel at animating scenes with high visual fidelity and maintaining the initial scene's identity, their primary design objective has been aesthetic realism and adherence to user-specified motion (often via text prompt), rather than physical plausibility. The evaluation of I2V models is typically focused on metrics like Fréchet Inception Distance (FID), Inception Score (IS), and temporal coherence measures, which assess visual quality and smoothness but do not inherently check for adherence to physical laws like gravity, friction, or object dynamics. This gap underscores the need for benchmarks like WorldBench, which can systematically test whether the generated motion reflects a true understanding of the physical parameters required for real-world simulation.

\subsection{Physics Datasets and Benchmarks}
There has been a growing interest in the community to evaluate the physical understanding and reasoning abilities of modern vision models~\cite{dasari2019robonet,groth2018shapestacks,piloto2018probing,riochet2018intphys,smith2019modeling, Kadambi2023-tv}. An early dataset was PHYRE~\cite{bakhtin2019phyre} which focuses on simple 2D scenarios constructed from balls and rectangular bars, with dynamics like collision, gravity, and friction. CLEVRER~\cite{yi2019clevrer} brings evaluation to 3D but is focused on the VQA task. More recently, the Physion dataset~\cite{bear2021physion} compiles a set of visually realistic videos testing various physics concepts. However, it uses object contact prediction (OCP), a binary task, to evaluate the physical understanding which limits its use for guiding development. While considerable progress has been made in this space, all prior work are deficient in at least one key area. Datasets like PHYRE and CLEVRER lack in visual realism and are made up of overly simplistic objects and structures. IntPhys2~\cite{bordes2025intphys} similarly uses a binary prediction task where a model must pick between a realistic and unrealistic video. Recent works Morpheus~\cite{zhang2025morpheus} and~\cite{motamed2026generative} works similarly to ours in terms of testing real physics scenarios, but outputs a "score" rather than exact constants, making it harder to compare and understand results. PISA-Bench~\cite{li2025pisa} works similar to our intuitive physics subset, but only tests videos of falling objects, limiting its scope greatly. Another recent line of work includes VideoPhy and similar papers~\cite{bansal2024videophy, bansal2025videophy, meng2024towards} which evaluate video models using VLMs as judges. While they can evaluate more varied videos and concepts, the use of VLMs means the metrics are not objective and can be hard to compare and reason about. Compared to these, our benchmark is the first benchmark where the inputs and outputs are both video based and where the output metrics are meaningful and objective. This aligns much more closely with the architectures of today's models, making it a better fit for physics evaluation.

\begin{figure*}
    \centering
    \includegraphics[width=0.9\linewidth]{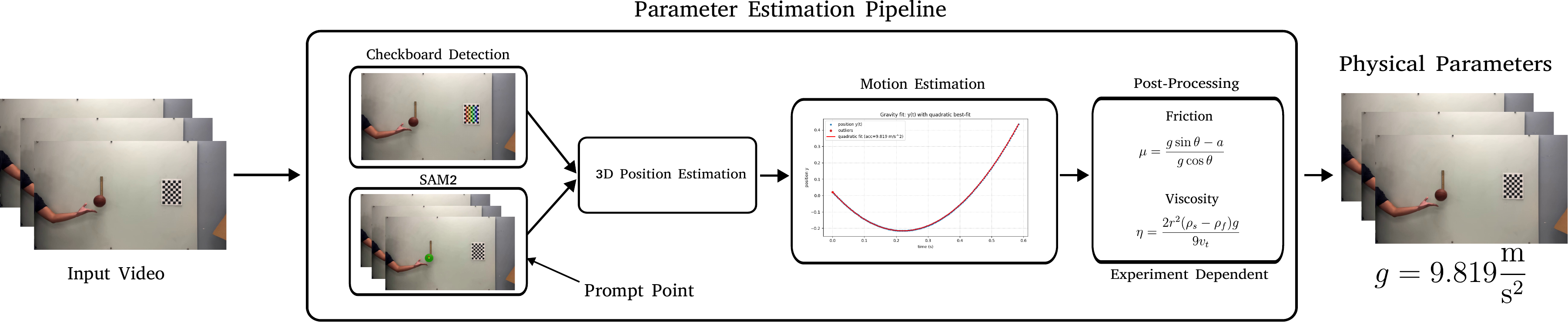}
    \caption{\textbf{Overview of our physical parameter estimation pipeline.} Given an input video, we first use checkerboard detection and SAM2 to extract 3D positions for objects in the video. We then fit curves to these parameters to estimate relevant physical properties such as acceleration or terminal velocity. These are then post-processed, if needed, to calculate the relevant physical parameters.}
    \label{fig:placeholder}
\end{figure*}

    

\begin{table}
  \caption{\textbf{Validation of Physics Parameters on the PerfectPhysics subset.} We show the evaluation pipeline on our captured ground truth videos here. All estimated parameters are within an acceptably small margin of error. Units for gravity are $m/s^2$ and units for Viscosity are Pascal-seconds (Pa.s). SP stands for Sandpaper.}
  \label{tab:perfphys_validation}
  \centering
  \setlength{\tabcolsep}{3pt} 
  \tiny
  \begin{tabular}{lcccccccccc}
    \toprule
      & \multicolumn{2}{c}{Gravity} & \multicolumn{3}{c}{Viscosity} & \multicolumn{5}{c}{Friction} \\
    Model & Free-Fall & Parabolic & Glycerine & Corn Syrup & Honey & Wood & Rubber & SP (80) & SP (3000) & Plastic \\
    \midrule
    Estimated & 9.78 $\pm$ 0.38 & 9.85 $\pm$ 0.36 & 1.22 $\pm$ 0.01 & 5.84 $\pm$ 0.02 & 13.82 $\pm$ 0.75 & 0.35 $\pm$ 0.05 & 0.93 $\pm$ 0.10 & 1.06 $\pm$ 0.05 & 0.30 $\pm$ 0.02 & 0.22 $\pm$ 0.03\\
    GT & 9.81 & 9.81 & 1.2 & 5.0 - 7.0 & 14.1  & 0.2 - 0.5 & 0.5 - 2.0 & 0.7 - 1.1 & 0.2 - 0.5 & 0.05 - 0.2 \\
    \bottomrule
  \end{tabular}
\end{table}

\begin{table*}
  \caption{\textbf{Foreground mIoU results on the Physical Principles Understanding subset (Simulated Videos).} Since the diffusion models generates 121 frames vs 33 for the autoregressive, we provide both comparisons. Higher is better for all columns}
  \label{tab:intuitivephys_miou_syn}
  \centering
  \tiny
  \begin{tabular}{lccccccccc}
    \toprule
    Model & Params & Ball Bounce &  2 Obj Fall &  2 Obj Para &  Block/Obj &  Columns &  Raised Block &  Walls &  Two Ball \\
    \midrule
    Cosmos-1 AR~\cite{agarwal2025cosmos} & 5B & \textbf{0.3759} & 0.2675 & 0.2268 & 0.2643 & 0.7032 & 0.3798 & 0.4996 & 0.1607 \\ 
    Cosmos-1 (33F)~\cite{agarwal2025cosmos} & 7B & 0.3719 & 0.2994 & \textbf{0.2831} & \textbf{0.3476} & 0.7349 & \textbf{0.4555} & \textbf{0.5578} & 0.2013 \\
    Cosmos-1 (121F)~\cite{agarwal2025cosmos} & 7B & 0.1636 & 0.1444 & 0.2008 & 0.2047 & 0.5575 & 0.3193 & 0.4155 & 0.1403 \\
    Cosmos-2~\cite{nvidia2025cosmospredict2} & 2B & 0.1903 & 0.1780 & 0.1401 & 0.1271 & 0.6958 & 0.4514 & 0.3298 & \textbf{0.2055} \\
    Cosmos-2~\cite{nvidia2025cosmospredict2} & 14B & 0.1913 & 0.2013 & 0.1351 & 0.1672 & 0.4758 & 0.1846 & 0.2699 & 0.1751\\
    Cosmos-2.5~\cite{ali2025world} & 2B & 0.1996 & \textbf{0.3035} & 0.1607 & 0.0992 & \textbf{0.7363} & 0.4480 & 0.3929 & \textbf{0.2055} \\
    \toprule
    & Params &  Obj Tow. &  Obj Away &  Sphere Tow. &  Sphere Away &  Dominoes &  Ramp &  Table  & Avg.\\
    \midrule
    Cosmos-1 AR~\cite{agarwal2025cosmos} & 5B & 0.2984 & 0.4121 & 0.6349 & 0.4799 & 0.4605 & \textbf{0.5292} & \textbf{0.6439} & 0.4225\\
    Cosmos-1 (33F)~\cite{agarwal2025cosmos} & 7B & \textbf{0.3272} & \textbf{0.4840} & \textbf{0.7123} & \textbf{0.5546} & \textbf{0.4892} & 0.4861 & 0.4573 & \textbf{0.4508} \\
    Cosmos-1 (121F)~\cite{agarwal2025cosmos} & 7B & 0.0996 & 0.2330 & 0.2453 & 0.1774 & 0.1568 & 0.3802 & 0.4215 & 0.2573\\
    Cosmos-2~\cite{nvidia2025cosmospredict2} & 2B & 0.1591 & 0.4341 & 0.4595 & 0.3337 & 0.2425 & 0.4082 & 0.4907 & 0.3201\\
    Cosmos-2~\cite{nvidia2025cosmospredict2} & 14B & 0.1484 & 0.3705 & 0.3721 & 0.3171 & 0.2067 & 0.4048 & 0.4061 & 0.2684\\
    Cosmos-2.5~\cite{ali2025world} & 2B & 0.1341 & 0.4321 & 0.4997 & 0.3189 & 0.2981 & 0.5049 & 0.4984 & 0.3488 \\
    \bottomrule
  \end{tabular}
\end{table*}

\section{Benchmark}

In order to test concept-specific physical understanding in "world foundation models" (WFMs), we introduce a novel benchmark, \bench, designed to evaluate their physics prediction capabilities. The core methodology is to provide these models with a short input video and tasking them to generate a continuation. Each video is designed to evaluate a single physics concept or law.

\bench~is divided into two distinct, yet complementary components, designed to probe both the intuitive and the engineering-grade understanding of the physical world. This structure ensures a comprehensive assessment of both high-level "intuitive physics" and low-level physics constants.

\subsection{Intuitive Physics Understanding}

The first subset is designed to assess the model's implicit understanding of core, foundational physics concepts, often referred to as ``intuitive physics." The goal is to determine if large-scale models, trained on vast quantities of video data, have successfully internalized these necessary cognitive building blocks. Our benchmark focuses on four fundamental physics concepts: Motion Physics (how objects move and interact), Support Relations (how objects are supported or balanced), Object Permanence (understanding that objects continue to exist when hidden), and Scale/Perspective (how size and spatial relationships change with motion/viewpoint). This is not an exhaustive list of physical concepts, but is designed to cover a range of common real-world scenarios. 

The difficulty in the design of this subset is narrowing the object trajectories to ensure consistent scene evolution, but allowing for enough variation for a rich and robust benchmark. To that end, for each concept, we construct 3-5 scenarios, where each of these scenarios is hand-designed to capture some element of the concept it is testing. Each scenario has 25 videos, each of which is generated by randomizing various components such as object type, location and material. In addition, we collect 10-14 real videos for each high-level concept, generally from a subset of the scenarios. In total, this subset of \bench~is made up of 469 videos spanning the 4 concepts: 425 simulated, and 44 real. Each video is 132 frames long and includes ground truth object segmentations. The synthetic videos additionally include ground truth depth, normals, and optical flow. All meshes and objects used in our simulations were taken from the ShapeNet dataset~\cite{chang2015shapenet} which includes 51,000 object models across 55 different categories. We sampled across all different categories and models, allowing for diversity in object shapes, textures, sizes, and properties.

We will now provide a brief description on each of the concepts. More details on individual scenarios along with sample images from the videos are included in the appendix in Section~\ref{sec:app_benchmarkdetails} and Section~\ref{sec:additional_results}.

\begin{itemize}
    \item \textbf{Motion Physics} is focused on evaluating the kinematics and dynamics in the generated video, specifically accounting for forces such as gravity and friction. This is a common real world scenario, as it is common for these models to have to simulate moving and colliding objects. To test motion physics, we create 3 scenarios: bouncing ball, two object fall, and two object parabolic motion.
    \item \textbf{Object Permanence} evaluates whether video generative models understand that objects continue to exist in the scenes even when hidden from the camera. This is a fundamental physics property that significantly affects our ability to predict the world. To test object permanence, we create 5 scenarios: block \& object, columns, raised block bounce, wall bouncing, and two ball bounce.
    \item \textbf{Support Relations} evaluates how objects physically support one another, e.g. one object preventing another from falling due to gravity or external forces. This includes understanding when certain configurations of objects are stable vs. unstable. To test support relations, we designed 3 scenarios: dominoes, ramp block, and table drop.
    \item \textbf{Perspective / Scale Relations} is designed to evaluate the accuracy of objects' appearance, such as size and location, with respect to the camera viewpoint. To evaluate perspective / scale relations, we designed 2 scenarios: obj/sphere moving towards camera and obj/sphere moving away from camera.
\end{itemize}

\subsubsection{Evaluation Methodology}

Our evaluation for the intuitive physics subset is done by comparing ground truth object segmentations with segmentations obtained from the generated videos. Specifically, our pipeline (visualized in Fig.~\ref{fig:overview}) works as follows: For every scenario, we use the ground truth segmentations to obtain bounding boxes of all objects in the video in the first frame. We then prompt SAM2~\cite{ravi2024sam2} with these bounding boxes, and use SAM2 to propogate these boxes/objects through the rest of the video. For every frame, we then compare the ground truth and predicted masks using the mIoU metric. We additionally, use the background region (calculated as all pixels not part of individual object masks) to compute the background RMSE. Since all of our backgrounds remain constant throughout the video, this metric measures how well the models maintain backgrounds.

\subsubsection{User Study}

In addition to directly obtaining the quantiative results in terms of IoU, we also perform a user study in order to see how well our IoU metrics are correlated with human judgements of physics accuracy. To do this, we recruited 50 participants and asked them to rate each video on a scale of 1-10 on four axes: Realism, Quality of Motion, Consistency, and overall Quality. Each user was shown 60 videos sampled evenly from all models. We then fit a line between IoU and user ratings and calculate the correlation coefficient. We provide a summary of the results here. More details about the study and results are provided in the appendix Section~\ref{sec:app_userstudy}. We find that across the four axes: \textbf{Realism: $r=0.768$, Quality of Motion: $r=0.846$, Consistency: $r=0.662$, Overall Quality: $r=0.713$.} Across all four categories, we show a consistently high correlation between our metrics and human judgements. This helps support the value of our metrics, as they can be obtained quickly and without human input while still being predictive of much more costly and tedious human study results.

\subsection{Physical Parameter Estimation}
The second subset shifts the focus from core physics principles to exact parameter estimation. One key goal of WFMs like NVIDIA's Cosmos~\cite{agarwal2025cosmos} is replacing physics simulation software (where parameters are hard-coded or manually tuned) as a synthetic data generator. To that end, it is crucial that these models are able to generate videos with accurate estimations for key physical parameters, such as gravitational acceleration. For this subset, we carefully designed a total of three experimental setups testing gravitational acceleration, friction coefficients, and fluid viscosity as the respective physical parameters. These concepts have 51 videos, 103 videos, and 80 videos respectively, for a total of 234 videos. We additionally synthetically generate 30 gravity and 15 friction videos, bring the combined total to 279 videos for the physical parameter estimation set.

The difficulty of designing these settings lies in reducing the effect of confounding variables that can influence the physical parameter estimation ability of these models (e.g. depth ambiguity, object motion uncertainty, etc.). To address this, each experimental setup and evaluation pipeline is meticulously designed to ensure a consistent evaluation. This setup and pipeline are detailed in Sec.~\ref{sec:physparams_eval}. We validate this pipeline by running it on our collected videos, and ensuring that the output physical parameter is close to the ground truth (e.g. $9.8 \frac{m}{s}^2$ for gravitational acceleration). Validation results can be seen in~\cref{tab:perfphys_validation}. 

We will now provide detail for the experimental setups, with additional information in the appendix Section~\ref{sec:app_benchmarkdetails}:

\begin{itemize}
    \item \textbf{Gravity} consists of 51 videos, 17 for straight drops and 34 for parabolic motion. Dropped items vary in shape and size. For parabolic motion, launch angle and trajectory vary as well.
    \item \textbf{Viscosity} consists of 80 videos, 32 for glycerine, 30 for corn syrup, and 18 for honey. In these videos, a steel ball is dropped into a test tube. The terminal velocity is estimated from the 3d position of the steel ball, which in turn gives us the viscosity. Note that viscosity varies depending on temperature and water absorption. As such, we collect all videos at $75\degree F$ and within the same session to as not to influence the viscosity of hygroscopic fluids.
    \item \textbf{Friction} consists of 103 videos, 30 for wood, 19 for rubber, 18 for sandpaper (80 grit), 18 for sandpaper (3000 grit), and 18 for plastic. In these videos, a steel block is dropped down a ramp covered with different materials. The angle of the ramp varies between runs.
\end{itemize}

\begin{table*}
  \caption{\textbf{Estimated Physics Parameters on the Physical Parameter Estimation subset (Real Videos).} Both ground truth ranges as well as our real video estimations are provided as comparison (note that our video estimations are all very close to or within the range of acceptable values). For gravity and viscosity, models closest to the ground truth values are bolded. Due to the wide range of acceptable friction values, models closest to our real video estimations are bolded. Units for Gravity are $m/s^2$ and units for viscosity are Pascal seconds (Pa.s). SP stands for Sandpaper.}
  \label{tab:perfphys_real}
  \centering
  \tiny
  \setlength{\tabcolsep}{2pt} 
  \resizebox{\textwidth}{!}{
  \begin{tabular}{lcccccccccc}
    \toprule
     & \multicolumn{2}{c}{Gravity} & \multicolumn{3}{c}{Viscosity} & \multicolumn{5}{c}{Friction} \\
    Model & Free-Fall & Parabolic & Glycerine & Corn Syrup & Honey & Wood & Rubber & SP (80) & SP (3000) & Plastic \\
    \midrule
    Ground Truth & 9.81 & 9.81 & 1.2 & 6.0 & 14.1 & 0.2-0.5 & 0.5-2.0 & 0.7-1.1 & 0.2-0.5 & 0.05-0.2 \\
    Real Videos & 9.78$\pm$0.38 & 9.85$\pm$0.36 & 1.22$\pm$0.01 & 5.84$\pm$0.02 & 13.82$\pm$0.75 & 0.35$\pm$0.05 & 0.93$\pm$0.10 & 1.06$\pm$0.05 & 0.30$\pm$0.02 & 0.22$\pm$0.03\\
    Cosmos 1 AR~\cite{agarwal2025cosmos} & 4.215$\pm$3.713 & 4.297$\pm$1.294 & 7.8$\pm$1.04 & \textbf{8.44}$\pm$2.11 & $>50$ & \textbf{0.541}$\pm$0.124 & 1.237$\pm$0.142 & 1.277$\pm$0.185 & 0.528$\pm$0.079 & 0.508$\pm$0.101 \\ 
    Cosmos 1 DM~\cite{agarwal2025cosmos} & 3.506$\pm$1.912 & 7.652$\pm$2.927 & 0.603$\pm$1.191 & 1.538$\pm$1.478 & 0.17$\pm$0.171 & 0.674$\pm$0.165 & 1.295$\pm$0.214 & 1.453$\pm$0.257 & \textbf{0.522}$\pm$0.139 & 0.481$\pm$0.119 \\
    Cosmos2-2B\cite{nvidia2025cosmospredict2} & \textbf{8.927}$\pm$4.791 & 8.228$\pm$3.813 & 0.252$\pm$0.394 & 1.091$\pm$0.309 & \textbf{9.845}$\pm$6.198 & 0.696$\pm$0.094 & 1.468$\pm$0.26 & 1.262$\pm$0.259 & 0.598$\pm$0.111 & 0.614$\pm$0.131 \\
    Cosmos2-14B\cite{nvidia2025cosmospredict2} & 8.428$\pm$3.478 & \textbf{9.145}$\pm$4.171 & 0.221$\pm$0.101 & 1.304$\pm$0.844 & 1.805$\pm$1.813 & 0.659$\pm$0.162 & \textbf{1.224}$\pm$0.410 & \textbf{1.144}$\pm$0.473 & 0.556$\pm$0.088 & \textbf{0.461}$\pm$0.393 \\
    Cosmos 2.5~\cite{ali2025world} & 4.778$\pm$4.474 & 5.375$\pm$2.763 & \textbf{0.802}$\pm$0.478 & 1.091$\pm$0.309 & 1.678$\pm$0.079 & 0.695$\pm$0.185 & 1.449$\pm$0.339 & 1.522$\pm$0.278 & 0.611$\pm$0.119 & 0.628$\pm$0.135 \\
    \midrule
    Wan 2.2~\cite{wan2025wan} & 0.378$\pm$0.784 & 1.682$\pm$3.065 & 3.418$\pm$5.119 & 3.162$\pm$3.169 & 3.281$\pm$2.758 & 0.668$\pm$0.101 & 1.321$\pm$0.223 & 1.476$\pm$0.242 & 0.516$\pm$0.155 & 0.576$\pm$0.105 \\
    Hunyuan~\cite{hunyuanworld2025tencent} & 0.369$\pm$0.788 & 0.206$\pm$0.407 & 7.203$\pm$3.709 & $>50$ & $>50$ & 0.662$\pm$0.086 & 1.432$\pm$0.227 & 1.049$\pm$0.222 & 0.531$\pm$0.076 & 0.569$\pm$0.104  \\
    CogVideoX~\cite{yang2024cogvideox} & -0.039$\pm$0.136 & 0.181$\pm$0.239 & 3.286$\pm$2.854 & 2.247$\pm$2.769 & 2.672$\pm$1.909 & 0.634$\pm$0.126 & 1.474$\pm$0.332 & 1.288$\pm$0.498 & 0.516$\pm$0.134 & 0.502$\pm$0.201 \\
    Runway 4.5~\cite{runwaygen45} & 1.478$\pm$2.711 & 3.551$\pm$1.118 & 1.955$\pm$0.773 & 3.367$\pm$1.115 & 2.871$\pm$0.693 & 0.913$\pm$0.402 & 2.319$\pm$0.933 & 0.303$\pm$0.519 & 0.786$\pm$0.252 & 0.613$\pm$0.284 \\
    Kling 3.0~\cite{team2025kling} & 4.461$\pm$1.316 & 2.917$\pm$0.771 & 2.568$\pm$0.872 & 1.947$\pm$1.573 & 1.343$\pm$0.988 & 0.748$\pm$0.972 & 1.883$\pm$0.673 & 1.259$\pm$1.044 & 1.779$\pm$0.818 & 0.443$\pm$1.437 \\
    LTX-2.0~\cite{ltx22026lightricks} & 0.478$\pm$0.971 & 3.771$\pm$0.735 & 0.971$\pm$0.347 & 3.447$\pm$1.417 & 3.369$\pm$2.778 & 0.416$\pm$1.669 & 0.337$\pm$3.794 & 0.224$\pm$0.571 & 0.267$\pm$0.697 & 1.479$\pm$1.671 \\
    \bottomrule
  \end{tabular}
  }
\end{table*}

\begin{table*}
  \caption{\textbf{Estimated Physics Parameters on the Physical Parameter Estimation subset (Simulated Videos).} Various different materials are tested for friction. We give the average RMSE error instead. Values closest to the ground truth value or with the lowest error are bolded.}
  \label{tab:perfphys_syn}
  \centering
  \footnotesize
  \begin{tabular}{lcccccccccc}
    \toprule
    & & \multicolumn{2}{c}{Gravity} & \multicolumn{1}{c}{Friction} \\
    Model & Params & Free-Fall ($m/s^2)$ &  Parabolic ($m/s^2)$ & Varied Materials Average Error ($\downarrow$)\\
    \midrule
    Ground Truth & & 9.81 & 9.81 & -- \\
    Cosmos-1 AR~\cite{agarwal2025cosmos} & 5B & 10.831 $\pm$ 2.925 & 7.408 $\pm$ 1.15 & 0.298 \\ 
    Cosmos-1 DM~\cite{agarwal2025cosmos} & 7B &7.530 $\pm$ 3.672 & 6.274 $\pm$ 1.595 & 0.231 \\
    Cosmos-2~\cite{nvidia2025cosmospredict2} & 2B & 4.288 $\pm$ 5.061 & 13.229 $\pm$ 4.312 & 0.232 \\
    Cosmos-2~\cite{nvidia2025cosmospredict2} & 14B & 4.230 $\pm$ 6.740 & 2.930 $\pm$ 3.023 & 0.274 \\
    Cosmos-2.5~\cite{ali2025world} & 2B & 10.494 $\pm$ 3.934 & 12.540 $\pm$ 3.296 & 0.217 \\
    \midrule
    Wan 2.2~\cite{wan2025wan} & 5B & 1.157 $\pm$ 2.631 & 0.062 $\pm$ 1.034 & 0.253 \\
    Hunyuan Video~\cite{hunyuanworld2025tencent} & 13B & 0.939 $\pm$ 0.719 & 0.677 $\pm$ 0.569 & 0.2275 \\
    CogVideoX~\cite{yang2024cogvideox} & 5B & -0.226 $\pm$ 2.392 & 0.309 $\pm$ 1.631 & 0.361 \\
    Runway Gen-4.5~\cite{runwaygen45} & N/A & 1.229 $\pm$ 1.165 & 3.119 $\pm$ 2.553 & 0.317\\
    Kling 3.0~\cite{team2025kling} & N/A & 5.837 $\pm$ 2.551 & 4.599 $\pm$ 3.671 & 0.436 \\
    LTX-2.0~\cite{ltx22026lightricks} & 14B & 0.568 $\pm$ 0.347 & 2.785 $\pm$ 2.426 & 0.243 \\
    \bottomrule
  \end{tabular}
\end{table*}

\subsubsection{Evaluation Methodology}
\label{sec:physparams_eval}
Extracting exact physical parameters from video is a difficult task because it requires estimating 3D positions of objects from only a monocular video. Estimating these 3D positions requires three key pieces of information: camera intrinsics/extrinsics, the 2D pixel location of the object and the depth of the object. 

To collect camera intrinsics, we calibrate the camera using a traditional checkerboard method prior to collecting data. To estimate the extrinsics dynamically for every video/setting, we place a checkerboard in the background of all of our videos. Since we know the 3D locations of the checkerboard corners, we can use them to estimate the camera extrinsics given the intrinsics. We additionally verify the camera intrinsics and extrinsics through manual measurement (e.g. we manually measure the distance between the camera and the checkerboard and compare). To extract the 2D locations of objects, we use SAM2~\cite{ravi2024sam2} prompted by a manually selected prompt point to track the object through the frames. We take the centroid of the object mask as its 2D pixel location. Robustly estimating the object's depth given just a monocular video is difficult, so we instead opt to design our setup so that the depth is always constant and can be measured exactly. We do this by placing our objects at a constant depth, and ensuring that they move only in a plane parallel to the camera plane.

Once we have estimated the 3D position of the object throughout the video, we can use those positions to estimate relevant physics parameters. To do this, we first estimate the acceleration of the object. This is done by fitting a quadratic equation to the object positions over time and taking the appropriate coefficients. For our gravity scenarios, we then use this acceleration directly and compare it to $g$. For the friction setups, we use the following equation to compute a friction coefficient from the acceleration:
$$\mu = \frac{g \sin{\theta} - a}{g \cos{\theta}}$$
where $\theta$ is the angle of the ramp and is pre-measured.

For viscosity, instead of estimating accleration, we estimate the terminal velocity of the object in the fluid. We do this by fitting a line to the object positions and taking the slope. We then use the following equation to estimate viscosity:
$$\eta = \frac{2 r^2 (\rho_s - \rho_f) g}{9 v_t}$$
where $\rho_s$ and $\rho_f$ are the densities of the sphere and fluid respectively and $v_t$ is the terminal velocity.

\subsubsection{Relative Ranking Evaluation}

In addition to comparing estimated parameters against the known ground truth values, we perform a ranking evaluation where we compute the Common Language Effect Size (CLES) for each model between different materials for friction and viscosity. Between two groups, the CLES provides the chance that a randomly sampled member from one group will be larger than a randomly sampled member from the other group. For example, when applied to our data it can tell us how often a model will give Sandpaper a higher friction coeff than Plastic and vice-versa, providing a more nuanced evaluation of the models. We provide detailed results in the appendix Section~\ref{sec:app_clesresults} for space reasons, but we provide a few takeaways here: 1. Models perform much worse on viscosity than on friction. Across all models, most of the viscosity comparisons come out to near chance showing the models are barely if at all differentiating and ranking the different mediums. However, on friction we see much more separation and see that for far apart pairs (e.g. Rubber vs Plastic most models perform very well). 2. Almost all the models across open source, closed source, I2V, V2V all show very similar performance. They tend to be confident on the same pairs (e.g. Plastic vs Rubber and Rubber vs SP3K) while also being unsure on the same pairs (e.g. Rubber vs SP80). This suggests that the large scale datasets all of these models train on are leading them to have similar physics understand and performance and raises interesting questions about convergent dynamics in multiple models.

\vspace{-3pt}

\begin{table*}
  \caption{\textbf{Foreground mIoU results on the Physical Principles Understanding subset (Real Videos).} Higher is better for all columns}
  \label{tab:intuitivephys_rmse_real}
  \centering
  \small
  \begin{tabular}{lcccccc}
    \toprule
    Model & Params & Motion Physics & Object Perm. & Scale & Support Relations & Avg. \\
    \midrule
    Cosmos-1 AR~\cite{agarwal2025cosmos} & 5B & \textbf{0.3826} & 0.3471 & \textbf{0.1634} & 0.7517 & \textbf{0.4112}\\ 
    Cosmos-1~\cite{agarwal2025cosmos} & 7B & 0.2156 & 0.3644 & 0.044 & 0.6362 & 0.3151  \\
    Cosmos-2~\cite{nvidia2025cosmospredict2} & 2B & 0.2931 & 0.3641 & 0.0910 & \textbf{0.7648} & 0.3783 \\
    Cosmos-2~\cite{nvidia2025cosmospredict2} & 14B & 0.2716 & \textbf{0.3697} & 0.0836 & 0.6521 & 0.3443 \\
    Cosmos-2.5~\cite{ali2025world} & 2B & 0.2896 & 0.3114 & 0.0905 & 0.6691 & 0.3402 \\
    \bottomrule
  \end{tabular}
\end{table*}
\section{Discussion}

We evaluate the Cosmos family of models and a number of image-to-video models on both subsets of \bench. Note that as of now, Cosmos is the only video-to-video WFM that is open-source, though we expect its introduction to expand the already growing research area. We evaluate Cosmos-1 (Auto-regressive and Diffusion), Cosmos-2 (2B and 14B) and Cosmos-2.5. For image-to-video models, we evaluate Wan 2.2, Hunyuan Video, and CogVideoX, Kling 3.0, Runway Gen-4.5 and Lightricks 2.0. Note that Kling and Runway are both closed source. 

Quantitative results from our evaluations are shown in Table~\ref{tab:intuitivephys_miou_syn}, Table~\ref{tab:intuitivephys_miou_real}, Table~\ref{tab:perfphys_real}, and Table~\ref{tab:perfphys_syn}. Results for the background RMSE are in the appendix Section~\ref{sec:additional_results}. We summarize some of our learnings about current WFMs below:

\vspace{-10pt}
\paragraph{Across the board, models both estimate parameters poorly and exhibit an extremely high variance in the parameter estimation subset.} Experiments captured in the physical parameter estimation subset contain highly constrained dynamic sequences in an effort to focus solely on the parameter being tested. Despite this, all models (both Cosmos-family and image-to-video) exhibit extremely high variance between rollouts. This is especially apparent in the gravity experiments where objects experience downward acceleration at varying levels, even between rollouts with the same object and object trajectory.

\vspace{-10pt}

\paragraph{Model outputs tend to follow realistic motion trajectories, while not adhering to realistic motion parameters.} For example, for gravity, all models tended to exhibit realistic motion paths most of the time (both parabolic trajectories and straight drops). However, they fail to abide by the proper $9.8 \frac{m}{s}^2$. Similar trends appear in the fluid viscosity and friction experiments. As remarked on before, this highlights the need for benchmarks that can target particular physical parameters, as visual realism alone is not sufficient for synthetic world data generators.
\vspace{-10pt}

\paragraph{Image-to-video models tended to under-perform on the gravitational acceleration subtask.} Image-to-video models suffer from lack of temporal information in the input, and so typically under-performed. However, this trend was exacerbated in the gravitational acceleration subtask, where lack of temporal information led to severely low or even negative gravity estimates.
\vspace{-10pt}

\paragraph{Models perform similarly on the synthetic and real versions of both benchmarks} We find that across both subsets, all tested models perform similar in synthetic and real scenarios. This suggests that the cause for poor performance is not the distribution gap between real world videos and synthetically generated test cases but rather due to poor physics understanding in the models. 
\vspace{-10pt}

\paragraph{Models do not handle material properties on the long-tail of real-world distributions well} We find that the models typically are unable to accurately simulate objects or materials which are far from average values such as Honey (very high viscosity relative) or Plastic (very low friction coefficient relative). Instead, they tend to simulate these materials closer to the average and do not differentiate well. Additionally, the variance does not meaningfully increase on these materials, showing that the model is consistent in these incorrect predictions.
\vspace{-10pt}

\paragraph{Models perform better on scenarios with longer object interaction periods.} In the intuitive physics subset, we find that for scenarios where objects interacts for longer, such as Ramp, Table, and Walls, perform much better than shorter interactions such as in Two Object Fall, Two Object Parabolic Motion, or Dominoes. We can also see this in the intuitive physics subset where the friction tests, in which an object is slowly sliding down a ramp, perform quite well compared to viscosity or gravity.
\vspace{-10pt}

\paragraph{Models tend to perform better on scenarios where strong training priors are present.} We find that across scenarios and subsets, models rely heavily on training priors for generating future frames. For example, in the intuitive physics subset, we find that the models handle balls rolling down the ramp very well, but have more trouble modeling the interaction of the ball and a block at the bottom of the ramp. This is likely because a ball rolling has been seen in training data, but obstructions on ramps are less likely. This suggests that these models are relying more on priors from training data for their predictions rather than understanding of physical parameters/laws.
\vspace{-10pt}

\paragraph{Limitations}
The goal of this benchmark is to disambiguate between different core physical concepts and parameters. As such, we would like to continue expanding the number of concepts that we test. As each setup requires careful experimental design and extensive validation (for the parameter estimation subset in particular), this is an ongoing process. We plan to add additional physical tasks (collision mechanics, optics, etc.). Additionally, while our video-based design leads to more nuanced and diagnostic evaluation, requiring visual inputs limits the number of models we can evaluate. We hope that as image/video-to-video models become more popular with the introduction of Cosmos, they can benefit from our benchmark.

\vspace{-5pt}
\section{Conclusion}
In this work, we introduce \bench, a new benchmark designed to evaluate concept-specific physical understanding in today's world-foundation models. Previous benchmarks assessed a model's visual realism through human-based metrics, or a model's dynamic trajectory adherence through coarse-grained metrics like object contact prediction. We leverage our benchmark's video-based and modular framework to directly measure adherence to physical concepts, constants, and material properties. This aids in our goal to 1) disambiguate between visually appealing and physically accurate generations, and 2) diagnose the core concept-specific failures of modern WFMs. We hope this benchmark can highlight the distinct challenges faced by current methods and can help guide the development and understanding of new ones as well.

{
\small
\bibliographystyle{plain}
\bibliography{main}
}


\appendix
\clearpage
\setcounter{page}{1}

\appendix

\section{Additional Benchmark Details}
\label{sec:app_benchmarkdetails}
We provide below more detailed descriptions of the benchmark, including scene descriptions, prompt choice, and experimental setups.

\subsection{Prompt Choice}
Regarding prompt selection, we test a variety of different prompts, providing varying degrees of scene and setup related detail. The final set of prompts were selected based on performance on the benchmark. The same prompts were used for each model that was tested. The prompts will be provided as part of the benchmark.

\subsection{Intuitive Physics Subset}

Here we provide a more detailed description of the scenes used in our benchmark:

\paragraph{Bouncing Ball} A sphere, initially at rest at some known height above the ground, is subject to fall freely under gravity. The provided video includes frames after the bounce, ensuring the model has the necessary information for prediction. In this scenario, the initial height of the ball and material (including bounciness) of the ball are randomized.

\paragraph{Two Object Fall}
Two distinct objects are initially at rest at different heights above the ground and at a slight horizontal offset. Both objects are released to freely fall under gravity, and then typically collide with each other and the ground. In this scenario, we randomize the shape and initial positions of both objects.  

\paragraph{Two Object Parabolic Motion} This scenario is a slight variation of the above scenario, where two distinct objects are placed on opposite sides of the scene. They are then projected towards each other at some randomized initial angle and projectile angle, not necessarily in the same vertical plane. In this scenario, the shapes, locations, and initial velocities of the objects are randomized to generate diverse scenes.

\paragraph{Block \& Obj} An object is moving from left to right behind a wall. The movement is linear and predictable, starting left of the wall with a randomized initial velocity. The object disappears behind the wall and reemerges on the right side. In this scenario, the type of object along with its initial velocity are randomized.

\paragraph{Columns} An object moving from left to right behind several thin columns. This is very similar to the Block \& Obj task, except that the object periodically disappears and reappears as it passes each column. The provided frames generally In this scenario, the type of object and initial velocity are randomized.

\paragraph{Raised Block Bounce} A sphere bouncing vertically behind a raised block. A sphere appears above the block when approaching its highest vertical position, and below the block when close to the ground, As it bounces, it is periodically occluded by the block and not visible to the camera. In this scenario, the mass, restitution (bounciness), and material of the sphere were randomized.

\paragraph{Wall Bouncing} A sphere rolling horizontally behind a block between two walls. As the sphere rolls from one side to another, it eventually collides with a wall, bounces off, and rolls back the other direction. The sphere is periodically occluded while behind the block, and reappears in a gap as it approaches a wall. In this scenario, the mass, initial velocity, and friction coefficient of the sphere were randomized. 

\paragraph{Two Ball Bounce} Two spheres bouncing vertically, with one larger sphere in front and one smaller sphere straight behind it. During this motion, the small sphere is periodically occluded by the large ball as their vertical positions diverge. In this scenario, the mass, restitution, and materials of both spheres were randomized.

\paragraph{Dominoes} This scenario features an object colliding with a series of standing dominoes on a surface. Depending on initial velocity, the object knocks over varying number of dominoes, which may subsequently topple onto one another. In this scenario, the type/shape of the object thrown and initial velocity are randomized.

\paragraph{Ramp Block} A sphere rolling down an incline until it encounters a fixed barrier at the end, causing it to stop. This tests two things, one whether the incline supports the ball as it rolls, and then if the block at the bottom supports and stops it. In this situation, the angle and length of the incline and the initial position of the sphere are randomized.

\paragraph{Table Drop} An object is positioned at a table's edge with a portion extending beyond the surface. This scenario directly challenges a model's understanding of the minimum conditions required for stable support and balance. Mere contact with a support surface is insufficient--the object requires adequate support distribution relative to its mass distribution. In this scenario, both the type/shape of the object and its location relative to the table were randomized. This also meant that physical parameters such as mass and restitution were in a sense randomized as they depended on the object chosen.

\paragraph{Obj/Sphere Moving Towards Camera} A single object (e.g. a sphere or miscellaneous irregular object) is launched from the background and moves towards the camera. As the object approaches, it should appear to increase in size due to perspective. In these scenarios, the type/shape of object and the initial velocity are randomized.

\paragraph{Obj/Sphere Moving Away From Camera} A single object (e.g. a sphere or miscellaneous irregular object) is launched from the background and moves towards the camera. As the object moves away, it should appear to decrease in size due to perspective. Similarly, in these scenarios, the type/shape of object and the initial velocity are randomized.

\subsection{Physical Parameter Estimation Subset}

\paragraph{Setup Details}
Scenes were all captured using an iPhone 13 slow-motion camera. A checkerboard pattern was placed in all scenes for camera intrinsic and extrinsic calibration. Camera poses were manually adjusted to ensure consistent and accurate pose for every scene capture. Depth is estimated through the checkerboard calibration, but also validated through manual measurement for each scene capture. 

Videos are trimmed and frame rate information is verified through metadata. SAM2 segmentation is used for 2D pixel location. Initial segmentation is provided through a prompt point (but optionally can use a language-guided prompt as well). Acceleration is estimated from 3D locations using a quadratic fit. For viscosity (which uses terminal velocity), a linear regression method is used to estimate the terminal velocity (videos are trimmed to ensure object is at terminal velocity at the first frame). All evaluation pipelines are validated before running them on video generation model outputs. This is done by testing them on our collected videos, and ensuring that parameter estimates line up with the generally accepted ground truth values (refer to Table  2 in the main paper).

A ruler is added on the free-fall, parabolic, and five friction setups to help video generation models get a sense of scale. Viscosity experiments include a beaker with markings to have a sense of scale. It is worth noting addtionally that video-to-video models receive enough input frames for the model to estimate the correct gravitational acceleration, viscosity, or friction coefficient.

\paragraph{Gravity}
For free-fall experiments, an object is dropped from rest and accelerates downward until it exits the frame. The video is trimmed so that the initial frame is already freely in the air. For parabolic motion experiments, objects are pushed up a ramp and fall towards the ground. The video is trimmed such that the object is already freely in the air in the initial frame. Ramp angle is randomized to ensure a variety. Objects for both experiment types are randomized and chosen to ensure the impact by air resistance is negligible.
\paragraph{Viscosity}
For the viscosity experiments, a large beaker containing either glycerine, corn syrup, honey is sitting upright on a table. A steel ball is dropped into the beaker and falls towards the bottom. The terminal velocity at which the ball flows through the liquid can determine the liquids viscosity. The video is trimmed to ensure that the object is already falling at terminal velocity in the initial frame. The size of steel ball is randomized to ensure a variety.
\paragraph{Friction}
For the friction experiments, a steel cube is sliding down a ramp covered with either wood, rubber, sandpaper (80 or 3000 grit), or plastic. Using the magnitude of the acceleration vector, we can determine the kinetic coefficient of friction of the ramp material. The video is trimmed to ensure that during the initial frame, there are no external forces other than gravity and friction acting on the object. The ramp angle is adjusted and randomized to ensure variety.

\section{Text-Enhanced Subset}
To supplement our image-to-video and video-to-video benchmarks, we additionally created a language-ased subset of the benchmark. This is to aid in the evaluation of modern vision-language models (VLMs) on physics understanding and prediction. This subset asks models to interpret visual details or predict physical outcomes, allowing us to assess their ability to do physical reasoning in diverse situations.

\begin{figure*}
    \centering
    \includegraphics[width=.8\textwidth]{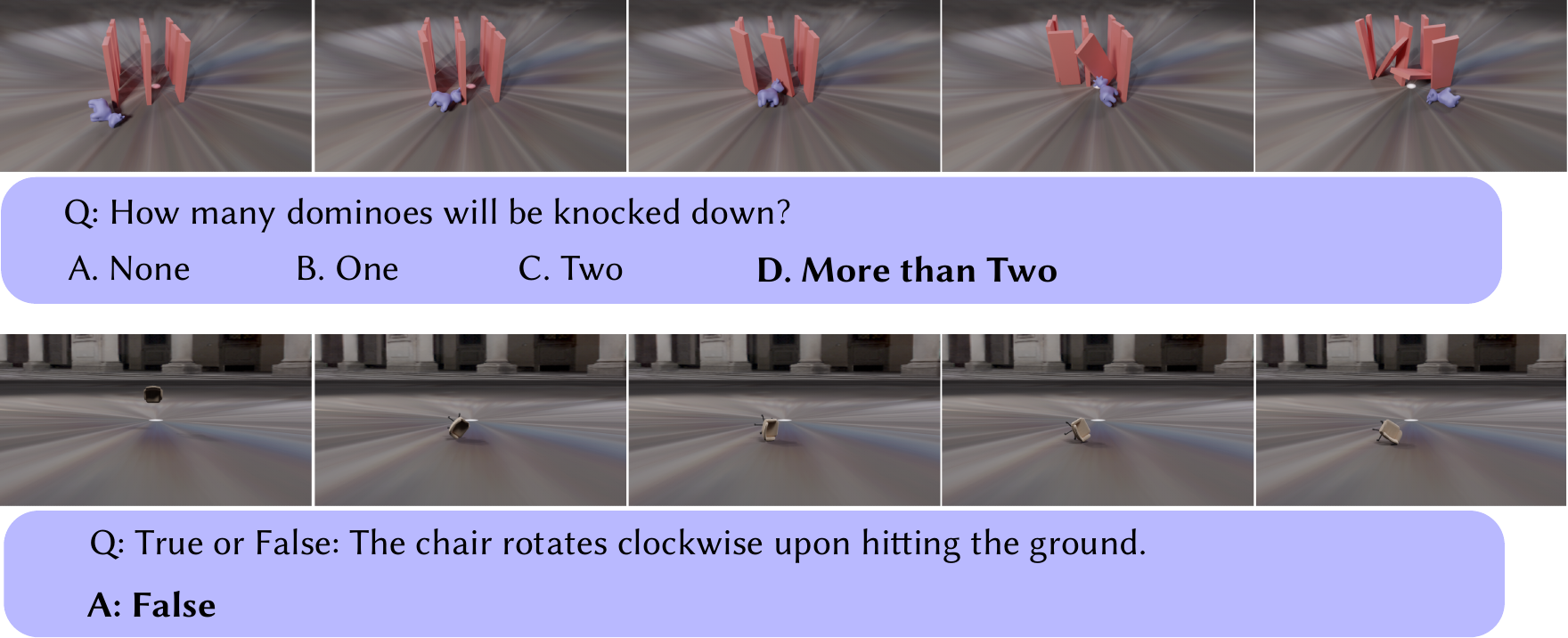}
    \caption{\textbf{Qualitative Examples of the Language-based subset of \bench}. VLMs are given access to a 9 frame video (same as what is inputted to COSMOS) and ask to answer a True/False or multiple choice question based on the video and future predictions.}
    \label{fig:language_qual}
\end{figure*}

We select a subset of 181 videos and write 1 natural language question per video. Questions can be either binary True/False questions or multiple choice with up to 4 choices. Example questions and answers are shown in Fig.~\ref{fig:language_qual}.

\subsection{Previous Work on Multi-modal VLMs}
Accompanying the wave of popularity of large language models is the vision-language model, a multi-modal model capable of processing both text and visual input~\cite{team2024gemini, wang2024qwen2, abouelenin2025phi, grattafiori2024llama, li2024llava, liu2023visual, yang2024thinking, zhu2023minigpt}. These models are typically capable of video understanding and reasoning. The recent Gemini model~\cite{team2023gemini, team2024gemini} is an example of a multi-modal model, capable of flexibly taking in any order of visual, textual, or audio input. The more recent Qwen2.5-VL model~\cite{wang2024qwen2, bai2025qwen2}, from the Qwen series of vision-language models, uses dynamic resolution processing and absolute time encoding, and particularly targets a visual agent's ability to perform visual reasoning, tool usage, and task execution. To evaluate the dynamics prediction accuracy and physical reasoning abilities of these models, we extend our proposed benchmark with a language-based visual reasoning framework.

\begin{figure*}[t]
    \centering
    \includegraphics[width=.95\textwidth]{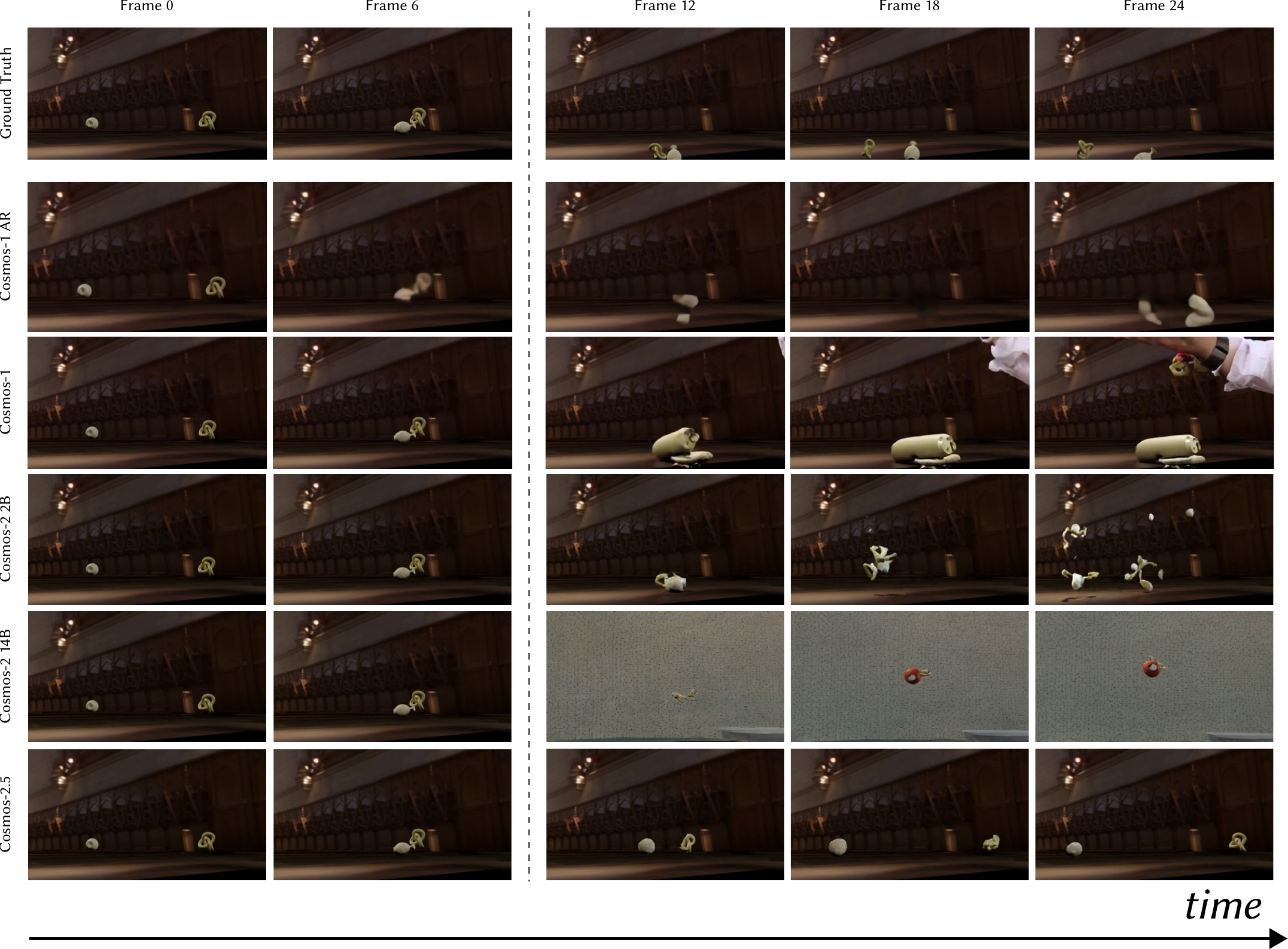}
    \caption{\textbf{Qualitative examples for the Motion Physics scenario of the intuitive physics subset.} For the motion physics example, two objects (a vase and a knot) are thrown at each other, collide, and then fall to the floor.}
    \label{fig:motion_phys_qual}
\end{figure*}

\begin{figure*}[t]
    \centering
    \includegraphics[width=.95\textwidth]{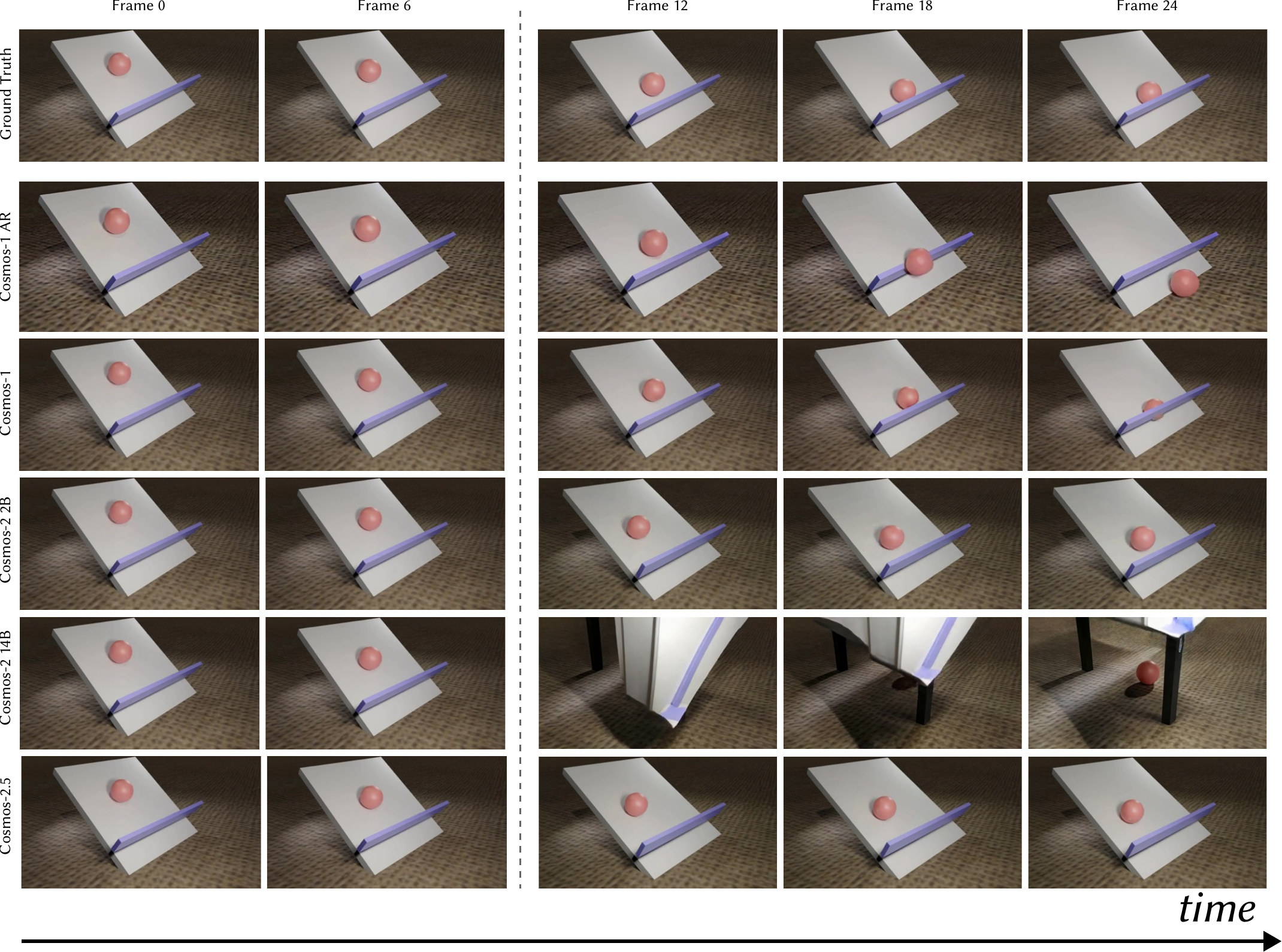}
    \caption{\textbf{Qualitative examples for the Support Relations scenario of the intuitive physics subset} A ball is rolled down a ramp towards a solid block near the bottom.}
    \label{fig:support_rel_qual}
\end{figure*}

\begin{figure*}[t]
    \centering
    \includegraphics[width=.95\textwidth]{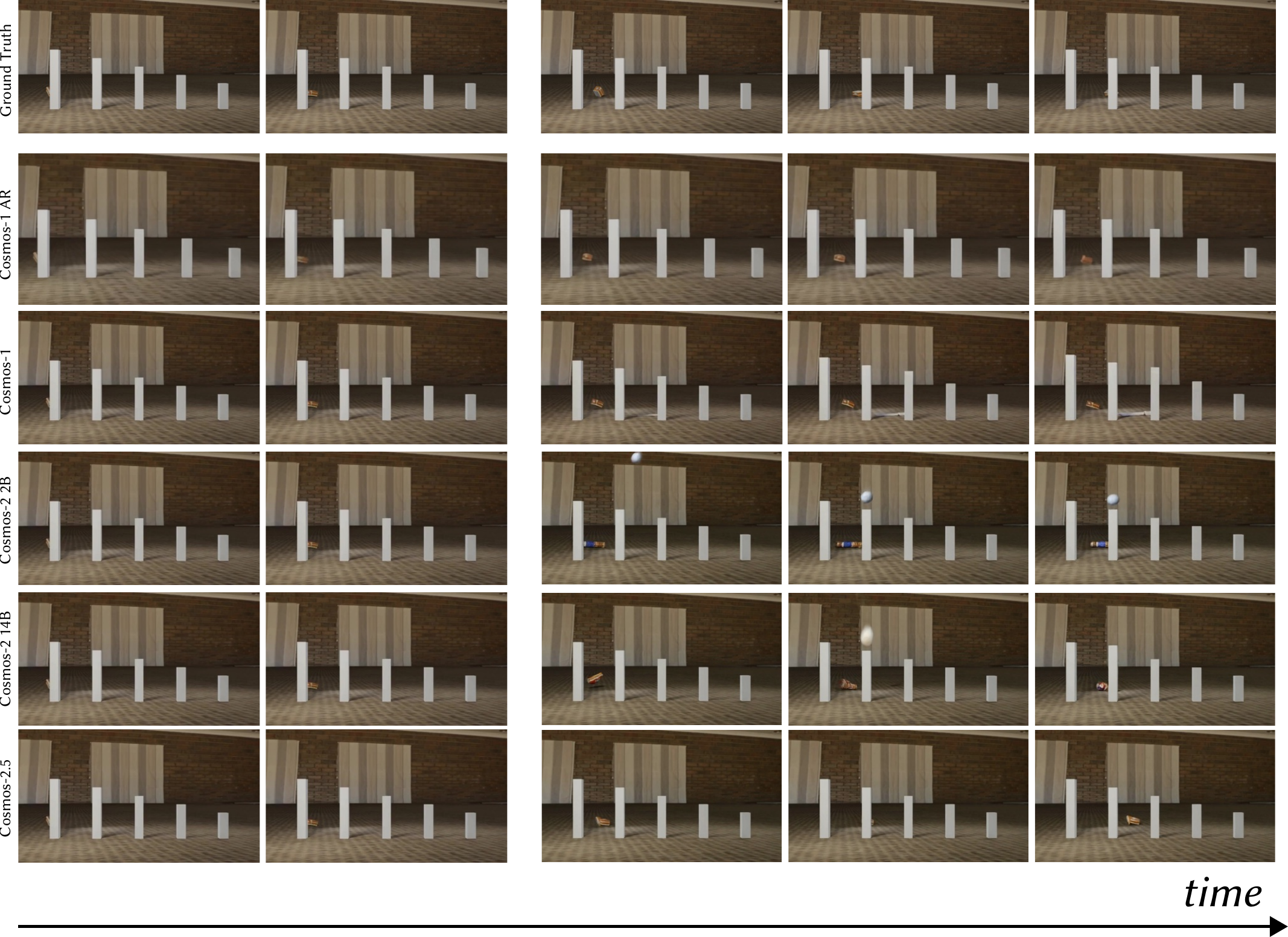}
    \caption{\textbf{Qualitative examples for the Object Permanence scenario of the intuitive physics subset}. An object is thrown behind a sequence of thin columns, appearing and dis-appearing as it goes.}
    \label{fig:obj_perm_qual}
\end{figure*}

\begin{figure*}[t]
    \centering
    \includegraphics[width=.95\textwidth]{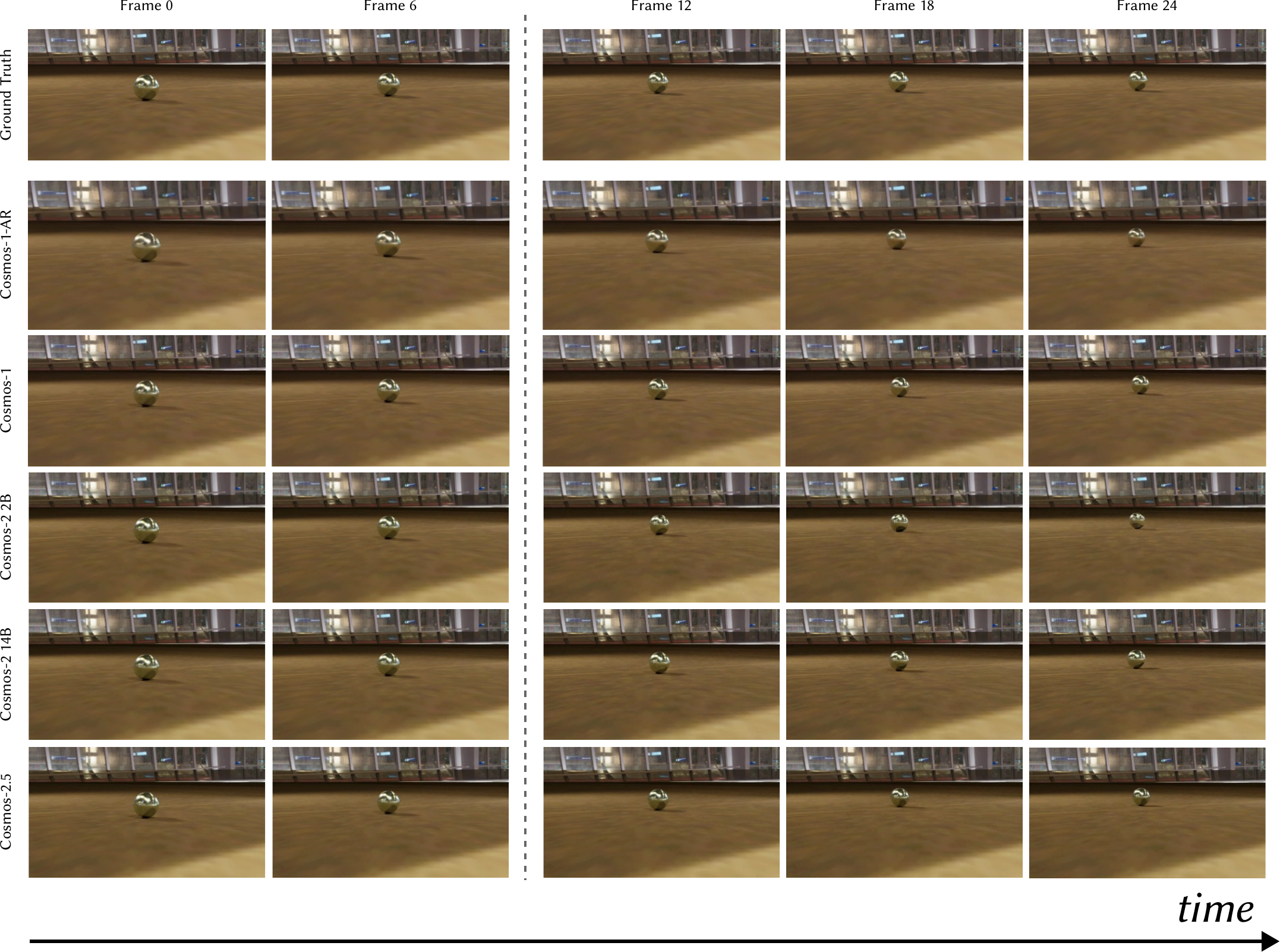}
    \caption{\textbf{Qualitative examples for the Scale/Perspective scenario of the intuitive physics subset.} A metallic sphere is rolling away from the camera. All models perform well on this sample.}
    \label{fig:scale_persp_qual}
\end{figure*}

\begin{figure*}[t]
    \centering
    \includegraphics[width=\textwidth]{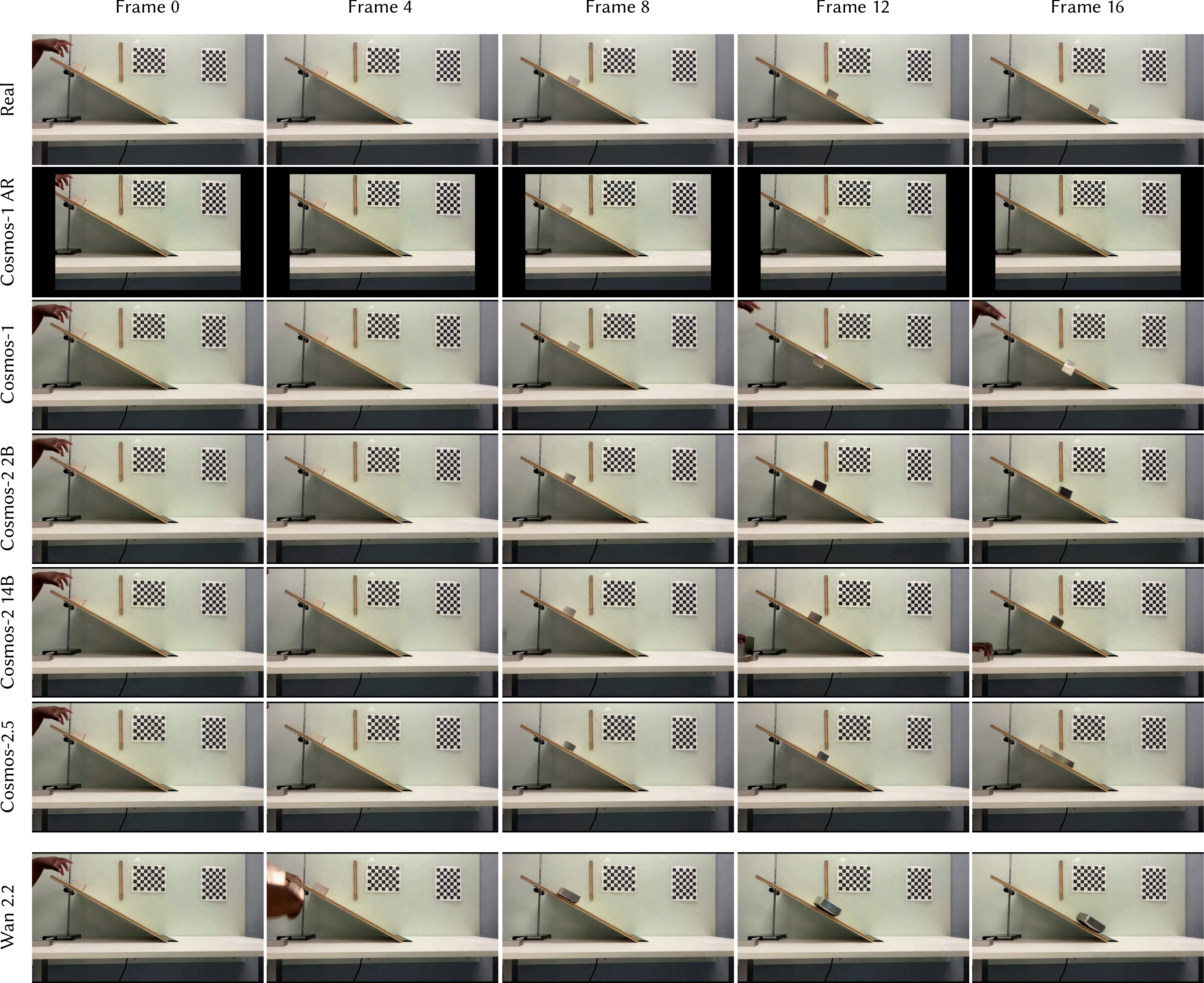}
    \caption{\textbf{Qualitative examples for the Friction scenario of the physical parameter estimation subset.} A steel block is released from rest down a ramp at a pre-set elevation and with a specific friction coefficient.}
    \label{fig:friction_qual}
\end{figure*}

\begin{figure*}[t]
    \centering
    \includegraphics[width=\textwidth]{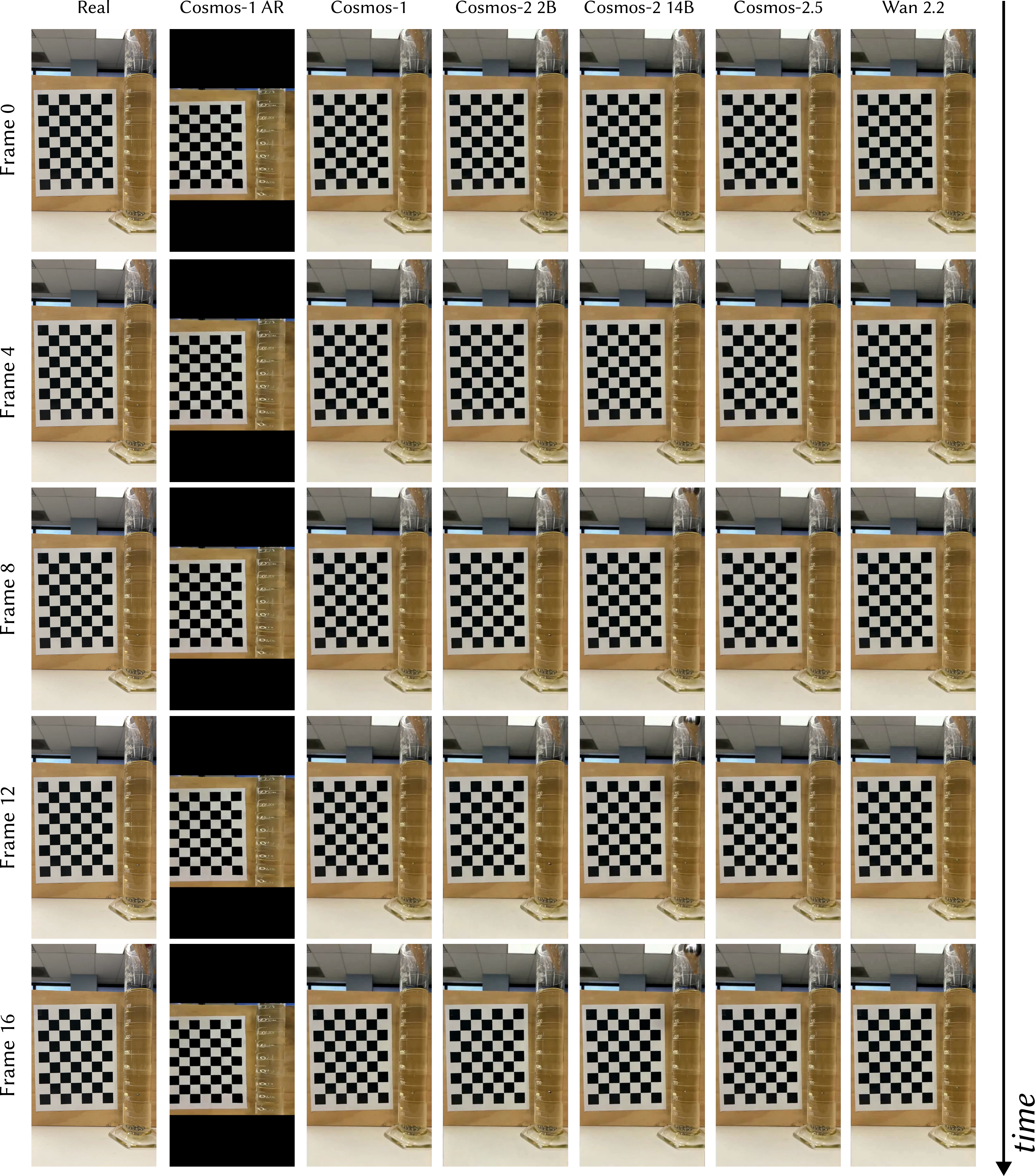}
    \caption{\textbf{Qualitative examples for the Motion Physics (top) and Scale/Perspective (bottom) scenarios.} For the motion physics example, two objects (a vase and a knot) are thrown at each other, collide, and then fall to the floor. The auto-regressive model greatly distorts the object shapes, while the diffusion model hallucinates the vase into a tank and adds a human hand. For the scale example, a metallic sphere is rolling away from the camera. Both models perform well on this sample.}
    \label{fig:viscosity_qual}
\end{figure*}

\subsection{Results}

In order to evaluate the language-based subset of \bench, we test SOTA closed- and open-source models: Qwen2.5 and Gemini. Qwen2.5 comes in 3 sizes, 7B, 32B, and 72B parameters, and is designed to handle vision inputs natively. For Gemini, we test both Gemini 2.5 Flash and Gemini 2.5 Pro. When running the evaluations, all models are provided with a system prompt which describes the tasks, defines what the output format should be, and the format of the data provided (9 frames). The Qwen models are run on 1 H100 GPU with the use of vLLM, while the Gemini models are evaluated through the provided API~\cite{kwon2023efficient}. The total costs for evaluation were approximately \$25.  The outputs of the models are evaluated by directly comparing against the answers. Results for all the models are shown in Table.~\ref{tab:vlm}. Across all 4 scenarios, Gemini 2.5 Pro performs the best overall, achieving 49.72\% accuracy. Within the open source models, Qwen2.5 32B surprisingly outperforms the larger 72B model, largely due to a very strong performance on the motion physics category. However, overall, all five models perform relatively poorly on the benchmark, achieving results only slightly better than chance. This suggests there is still much work to be done to improve the physical understanding of modern VLMs.

\begin{table*}[h]
\caption{\textbf{Results of SOTA closed and open models on our language-based benchmark}. Gemini 2.5 Pro, a closed model performs best overall, and Qwen2.5 32B performs best among open models.}
\label{tab:vlm}
\centering
\small
\begin{tabular}{ccccccc}
\toprule
 & Model &  Motion Phys & Obj. Perm. & Scale/Persp. & Support Rel. & \textbf{Avg.}  $\uparrow$  \\
\hline
\multirow{6}{*}{\shortstack{Open \\ Models}}
  & Qwen2.5-VL-7B~\cite{bai2025qwen2}   & 0.5161 & 0.2381 & 0.4474 & 0.5357 & 0.3737 \\
  & Qwen2.5-VL-32B~\cite{bai2025qwen2}  & \textbf{0.8710} & 0.2738 & 0.4737 & \textbf{0.5714} & 0.4641 \\
  & Qwen2.5-VL-72B~\cite{bai2025qwen2}  & 0.5806 & 0.3333 & 0.4211 & \textbf{0.5714} & 0.4309 \\
  & GLM 4.1V 9B~\cite{hong2025glm} & 0.6674 & 0.3453 & 0.4473 & 0.6071 & 0.4641 \\
  & Mistral Small 3.2 24B & 0.4838 & 0.2500 & 0.3684 & 0.3571 & 0.3315 \\
& Llama-3.2-11B-Vision & 0.5161 & 0.1548 & 0.3421 & 0.3571 & 0.2873 \\
\midrule
\multirow{4}{*}{\shortstack{Closed \\ Models}}
  & Gemini 2.5 Flash~\cite{team2024gemini} & 0.6452 & 0.3571 & \textbf{0.6053} & 0.4643 & 0.4751 \\
  & Gemini 2.5 Pro~\cite{team2024gemini}  & 0.6774 & 0.4048 & 0.5000 & \textbf{0.5714} & 0.4972 \\
  & Claude Sonnet 4 & 0.7096 & 0.4286 & 0.5526 & \textbf{0.4285} & \textbf{0.5027} \\
  & GPT 4.1 & 0.3781 & 0.2619 & 0.5000 & 0.5000 & 0.3701 \\
  \bottomrule
\end{tabular}
\end{table*}

\subsection{Additional Quantitative VLM Metrics}

In this section, we expand upon the results in Tab. 3 of the main paper and show the results for each model by scene category.  Although all 5 models tested perform similarly in most categories, we notice striking differences in the Walls category where the Qwen models are all near 0.0 while both Gemini models achieve accuracys $>$ 0.6. All models have the most trouble with the Object Permanence scenes and perform the best on Motion Physics scenes overall. This is somewhat expected as it is likely that their training data included more examples of motion physics than of object permanence.

\section{Additional Results}
\label{sec:additional_results}
In this section, we provide additional video generation results on our dataset, including qualitative results showing scene rollouts, background rmse results, as well as scene evolution over time.

\subsection{Additional Qualitative Results}
We provide here qualitative results of video generations on both subsets of our benchmark. For the intuitive physics subset, we show qualitative results in Fig.~\ref{fig:motion_phys_qual}, Fig.~\ref{fig:obj_perm_qual}, Fig.~\ref{fig:scale_persp_qual}, and Fig.~\ref{fig:support_rel_qual}. For the physical parameter estimation subset, we show results in Fig.~\ref{fig:friction_qual}, Fig.~\ref{fig:viscosity_qual}. Note that, as mentioned in the main paper, results varied greatly between generations, and results in one rollout as seen in the figures is not indicative of broad trends throughout the benchmark.

\subsection{Background RMSE Results}
For the intuitive physical understanding subset, we use the foreground mIoU metric to evaluate video generation performance. As an additional metric, we provide here background RMSE results. Results are in~\cref{tab:intuitivephys_rmse_syn} and ~\cref{tab:intuitivephys_rmse_real}

\subsection{Results Over Time}
\begin{figure*}
    \centering
    \includegraphics[width=\textwidth]{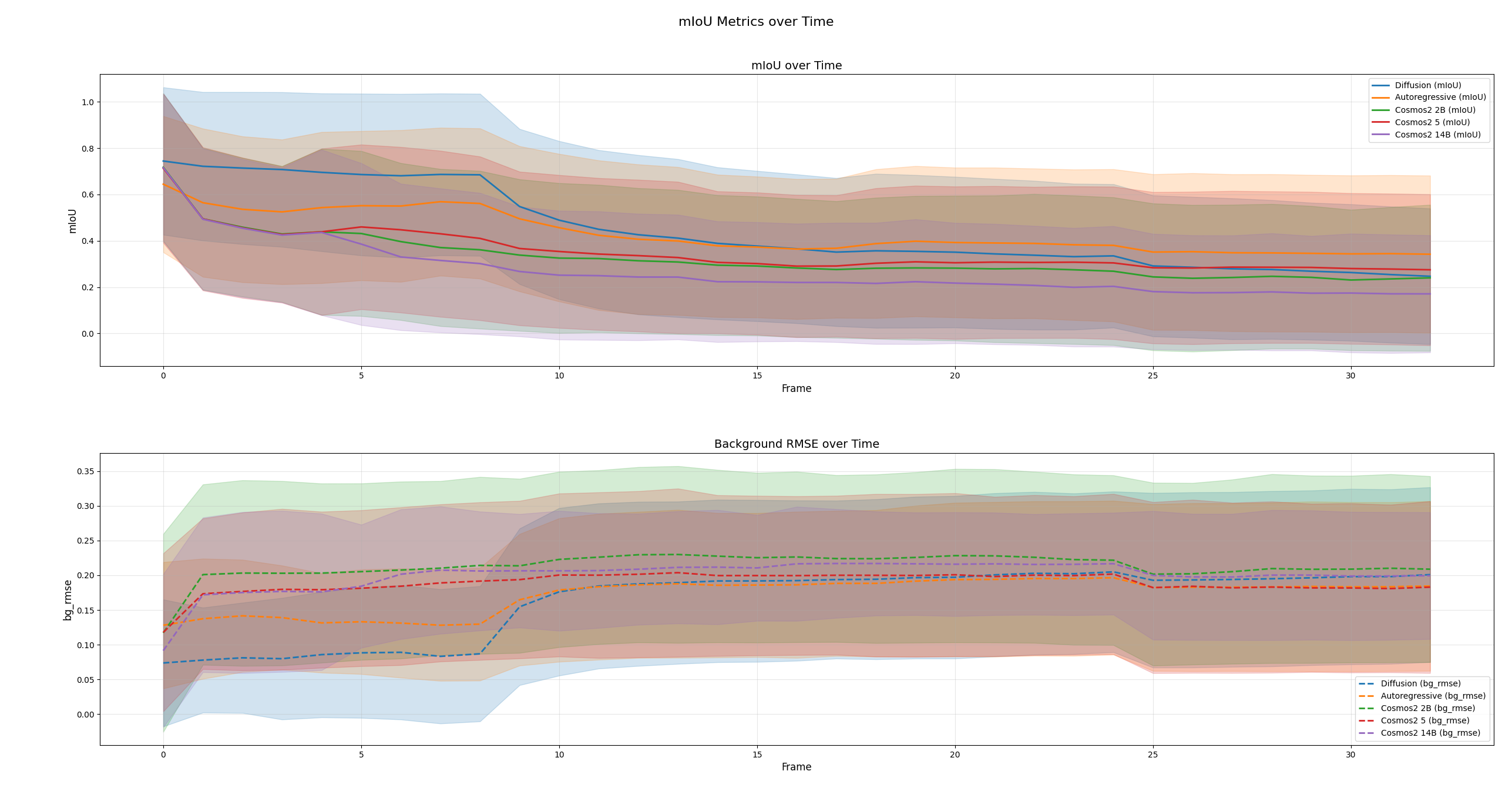}
    \caption{\textbf{mIoU and background RMSE results over time}. The foreground mIoU is inversely related with how far in the future the model is predicting. There is a sharp drop off after frame 5 or frame 9 when the model first begins predicting (depending on the model) The shaded region shows 1 standard deviation.}
    \label{fig:miou_time}
\end{figure*}

For the intuitive physical understanding subset, rollouts can be especially long (as opposed to the short physics experiments in the second subset). Here, we analyze the differences between the 30 frame and 120 frame rollout length. As seen in~\cref{fig:miou_time}, foreground mIoU is inversely related with how far in the future the model is predicting, suggesting a compounding error effect in physical accuracy. This highlights the need for accurate physical understanding, as compounding error effects can be especially detrimental if leveraging world models as synthetic data generators.

\begin{table*}
  \caption{\textbf{Background RMSE results for Cosmos models on our benchmark.} RMSE is computed over only the ground truth segmentation for "background". Lower is better for all columns.}
  \label{tab:intuitivephys_rmse_syn}
  \centering
  \tiny
  \begin{tabular}{cccccccccc}
    \toprule
    Model & Params &Ball Bounce &  Two Obj Fall &  Two Obj Para &  Block/Obj &  Columns &  Raised Block &  Walls &  Two Ball \\
    \midrule
    Cosmos-1 AR & 5B & \textbf{0.2112} & 0.2168 & 0.2542 & \textbf{0.1943} & 0.1376 & 0.0978 & 0.2299 & 0.1440 \\ 
    Cosmos-1 & 7B & 0.2403 & 0.2221 & 0.2594 & 0.2103 & 0.1104 & 0.1223 & 0.2223 & 0.1709 \\
    Cosmos-2 & 2B & 0.2459 & 0.2429 & 0.2370 & 0.1970 & \textbf{0.0799} & \textbf{0.0344} & 0.2212 & 0.1368 \\
    Cosmos-2 & 14B & 0.2265 & \textbf{0.2095} & \textbf{ 0.2177} & 0.2086 & 0.1401 & 0.1393 & \textbf{0.2027} & 0.1601 \\
    Cosmos-2.5 & 2B & 0.2586 & 0.2361 & 0.2659 & 0.2227 & 0.0853 & 0.0307 & 0.2117 & \textbf{0.1236} \\
    \toprule
    & Params & Obj Tow. &  Obj Away &  Sphere Tow. &  Sphere Away &  Dominoes &  Ramp &  Table  & Avg.\\
    \midrule
    Cosmos-1 AR & 5B & 0.2226 & \textbf{0.1696} & \textbf{0.1218} & \textbf{0.1848} & 0.1616 & \textbf{0.1027} & 0.1715 & \textbf{0.1747}\\
    Cosmos-1 & 7B & 0.2534 & 0.2329 & 0.1325 & 0.2256 & \textbf{0.0998} & 0.2283 & 0.2586 & 0.1993\\
    Cosmos-2 & 2B & 0.2551 & 0.2298 & 0.1760 & 0.1936 & 0.2231 & 0.3182 & 0.3222 & 0.2075 \\
    Cosmos-2 & 14B & \textbf{0.2108} & 0.2094 & 0.2061 & 0.1886 & 0.2015 & 0.1947 & 0.2382 & 0.1969 \\
    Cosmos2.5 & 2B & 0.2593 & 0.2355 & 0.1738 & 0.2021 & 0.1952 & 0.1488 & \textbf{0.1669} & 0.1877 \\
    \bottomrule
  \end{tabular}
\end{table*}

\begin{table*}
  \caption{\textbf{Background RMSE results on the Physical Principles Understanding subset (Real Videos).} Lower is better for all columns}
  \label{tab:intuitivephys_miou_real}
  \centering
  \small
  \begin{tabular}{ccccccc}
    \toprule
    Model & Params & Motion Physics & Object Perm. & Scale & Support Relations & Avg. \\
    \midrule
    Cosmos-1 AR~\cite{agarwal2025cosmos} & 5B & \textbf{0.0887} & 0.2432 & 0.2568 & \textbf{0.2295} & \textbf{0.2045}\\ 
    Cosmos-1~\cite{agarwal2025cosmos} & 7B & 0.1011 & 0.2100 & 0.2935 & 0.3090 & 0.2284  \\
    Cosmos-2~\cite{nvidia2025cosmospredict2} & 2B & 0.1589 & 0.2381 & 0.2681 & 0.2646 & 0.2324 \\
    Cosmos-2~\cite{nvidia2025cosmospredict2} & 14B & 0.1289 & \textbf{0.2015} & \textbf{0.2358} & 0.3018 & 0.2170 \\
    Cosmos-2.5~\cite{ali2025world} & 2B & 0.1626 & 0.2317 & 0.2386 & 0.2784 & 0.2278 \\
    \bottomrule
  \end{tabular}
\end{table*}

\section{User Study Details}
\label{sec:app_userstudy}
In this section, we provide further details on the user study. The user study was done using the Prolific website. There were 50 participants who each reviewed 60 videos. Each video was rated on a scale from 1-10. Participants took an average of 30 minutes to complete the task and were paid \$12/hr for the task. Participants were recruited from around the globe using the Prolific matching service. The user study was done using Google Forms. A total of 54 participants began the study with 3 not finishing and 1 submission being rejected for low quality (all videos ranked the same). Results were provided in the main paper but we reproduce them in Table~\ref{tab:userstudyresults}.

\begin{table}[]
    \caption{Correlation Coefficient between User Study ratings and mIoU values obtained from our evaluation.}
    \centering
    \begin{tabular}{cc}
    \toprule
        Category & Correlation Coefficient \\
        \midrule
        Realism & 0.768 \\
        Quality of Motion & 0.846 \\
        Consistency & 0.662 \\
        Overall Quality & 0.713 \\
        \bottomrule
    \end{tabular}
    \label{tab:userstudyresults}
\end{table}

\section{SAM Validation}
\label{sec:app_samvalidation}
\begin{figure*}
    \centering
    \includegraphics[width=\textwidth]{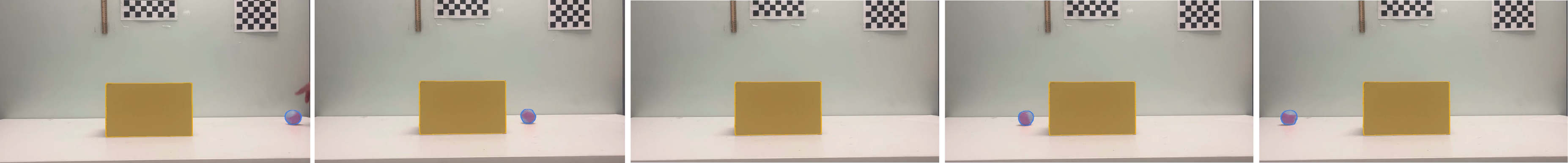}
    \caption{\textbf{SAM2 tracking an object in the object permanence scenario}. Despite the fact that the object disappears for 20+ frames, SAM2 is able to track it when it re-appears.}
    \label{fig:sam2_val}
\end{figure*}

For both our benchmark subsets, we rely on SAM2 for object segmentation and tracking, so we perform validation checks to evaluate SAM2 accuracy. Specifically, since we hand-annotate the real part of the intuitive physics subset, we compare those annotations against automatic SAM2 segmentations based on our prompting methods. We find an overall mIoU of 0.9445, showing a high overlap between hand-annotated and SAM2 segmentations. Additionally, in Fig.~\ref{fig:sam2_val}, we show SAM2 tracking an object in the object permanence scenario: even though the object disappears behind another object for 20+ frames, SAM2 continues to track it well due to prompting and maintained state.

\section{Common Language Effect Size Results}
\label{sec:app_clesresults}
In this section, we provide comprehensive results for the Common Language Effect Size metric computed on all models for both the friction and viscosity settings. Tables 9-24 contain per-model and per-scenario results.

Overall, we find that almost all models perform quite poorly on Viscosity, hovering around 50\% chance for almost all pairs. In terms of friction, we see more separation, with most models getting distant pairs (such as plastic vs sandpaper 80 grit) correct while having trouble with closer pairs.

\noindent
\begin{minipage}[t]{0.48\textwidth}
\centering
\small
\captionof{table}{CogVideo Friction}
\begin{tabular}{llrr}
\toprule
\textbf{Group X} & \textbf{Group Y} & \textbf{P(X>Y)} & \textbf{P(Y>X)} \\
\midrule
Plastic & SP 3K & 0.5791 & 0.4209 \\
Plastic & SP 80 & 0.1049 & 0.8951 \\
Plastic & Wood & 0.3074 & 0.6926 \\
SP 3K & SP 80 & 0.1090 & 0.8910 \\
SP 3K & Wood & 0.2385 & 0.7615 \\
SP 80 & Wood & 0.8889 & 0.1111 \\
\bottomrule
\end{tabular}
\end{minipage}
\hfill
\begin{minipage}[t]{0.48\textwidth}
\centering
\small
\captionof{table}{CogVideo Viscosity}
\begin{tabular}{llrr}
\toprule
\textbf{Group X} & \textbf{Group Y} & \textbf{P(X>Y)} & \textbf{P(Y>X)} \\
\midrule
Corn Syrup & Glycerine & 0.2474 & 0.7526 \\
Corn Syrup & Honey & 0.3185 & 0.6815 \\
Glycerine & Honey & 0.6325 & 0.3675 \\
\bottomrule
\end{tabular}
\end{minipage}

\vspace{2em}

\noindent
\begin{minipage}[t]{0.48\textwidth}
\centering
\small
\captionof{table}{Cosmos1-AR Friction}
\begin{tabular}{llrr}
\toprule
\textbf{Group X} & \textbf{Group Y} & \textbf{P(X>Y)} & \textbf{P(Y>X)} \\
\midrule
Plastic & Rubber & 0.0000 & 1.0000 \\
Plastic & SP 3K & 0.4300 & 0.5700 \\
Plastic & SP 80 & 0.0000 & 1.0000 \\
Plastic & Wood & 0.3704 & 0.6296 \\
Rubber & SP 3K & 1.0000 & 0.0000 \\
Rubber & SP 80 & 0.4649 & 0.5351 \\
Rubber & Wood & 1.0000 & 0.0000 \\
SP 3K & SP 80 & 0.0000 & 1.0000 \\
SP 3K & Wood & 0.4500 & 0.5500 \\
SP 80 & Wood & 1.0000 & 0.0000 \\
\bottomrule
\end{tabular}
\end{minipage}
\hfill
\begin{minipage}[t]{0.48\textwidth}
\centering
\small
\captionof{table}{Cosmos1-AR Viscosity}
\begin{tabular}{llrr}
\toprule
\textbf{Group X} & \textbf{Group Y} & \textbf{P(X>Y)} & \textbf{P(Y>X)} \\
\midrule
Corn Syrup & Glycerine & 0.5644 & 0.4356 \\
Corn Syrup & Honey & 0.5519 & 0.4481 \\
Glycerine & Honey & 0.5481 & 0.4519 \\
\bottomrule
\end{tabular}
\end{minipage}

\vspace{2em}

\noindent
\begin{minipage}[t]{0.48\textwidth}
\centering
\small
\captionof{table}{Cosmos1-Diffusion-7B Friction}
\begin{tabular}{llrr}
\toprule
\textbf{Group X} & \textbf{Group Y} & \textbf{P(X>Y)} & \textbf{P(Y>X)} \\
\midrule
Plastic & Rubber & 0.0000 & 1.0000 \\
Plastic & SP 3K & 0.4321 & 0.5679 \\
Plastic & SP 80 & 0.0000 & 1.0000 \\
Plastic & Wood & 0.1759 & 0.8241 \\
Rubber & SP 3K & 1.0000 & 0.0000 \\
Rubber & SP 80 & 0.3450 & 0.6550 \\
Rubber & Wood & 0.9772 & 0.0228 \\
SP 3K & SP 80 & 0.0000 & 1.0000 \\
SP 3K & Wood & 0.2426 & 0.7574 \\
SP 80 & Wood & 1.0000 & 0.0000 \\
\bottomrule
\end{tabular}
\end{minipage}
\hfill
\begin{minipage}[t]{0.48\textwidth}
\centering
\small
\captionof{table}{Cosmos1-Diffusion-7B Viscosity}
\begin{tabular}{llrr}
\toprule
\textbf{Group X} & \textbf{Group Y} & \textbf{P(X>Y)} & \textbf{P(Y>X)} \\
\midrule
Corn Syrup & Glycerine & 0.6012 & 0.3988 \\
Corn Syrup & Honey & 0.3089 & 0.6911 \\
Glycerine & Honey & 0.2667 & 0.7333 \\
\bottomrule
\end{tabular}
\end{minipage}

\vspace{2em}

\noindent
\begin{minipage}[t]{0.48\textwidth}
\centering
\small
\captionof{table}{Cosmos2-14B Friction}
\begin{tabular}{llrr}
\toprule
\textbf{Group X} & \textbf{Group Y} & \textbf{P(X>Y)} & \textbf{P(Y>X)} \\
\midrule
Plastic & Rubber & 0.0556 & 0.9444 \\
Plastic & SP 3K & 0.5401 & 0.4599 \\
Plastic & SP 80 & 0.1019 & 0.8981 \\
Plastic & Wood & 0.2815 & 0.7185 \\
Rubber & SP 3K & 0.9415 & 0.0585 \\
Rubber & SP 80 & 0.5351 & 0.4649 \\
Rubber & Wood & 0.9175 & 0.0825 \\
SP 3K & SP 80 & 0.0926 & 0.9074 \\
SP 3K & Wood & 0.2500 & 0.7500 \\
SP 80 & Wood & 0.8519 & 0.1481 \\
\bottomrule
\end{tabular}
\end{minipage}
\hfill
\begin{minipage}[t]{0.48\textwidth}
\centering
\small
\captionof{table}{Cosmos2-14B Viscosity}
\begin{tabular}{llrr}
\toprule
\textbf{Group X} & \textbf{Group Y} & \textbf{P(X>Y)} & \textbf{P(Y>X)} \\
\midrule
Corn Syrup & Glycerine & 0.6012 & 0.3988 \\
Corn Syrup & Honey & 0.4556 & 0.5444 \\
Glycerine & Honey & 0.3698 & 0.6302 \\
\bottomrule
\end{tabular}
\end{minipage}

\vspace{2em}

\noindent
\begin{minipage}[t]{0.48\textwidth}
\centering
\small
\captionof{table}{Cosmos2-2B Friction}
\begin{tabular}{llrr}
\toprule
\textbf{Group X} & \textbf{Group Y} & \textbf{P(X>Y)} & \textbf{P(Y>X)} \\
\midrule
Plastic & Rubber & 0.0000 & 1.0000 \\
Plastic & SP 3K & 0.5710 & 0.4290 \\
Plastic & SP 80 & 0.0370 & 0.9630 \\
Plastic & Wood & 0.3111 & 0.6889 \\
Rubber & SP 3K & 1.0000 & 0.0000 \\
Rubber & SP 80 & 0.6520 & 0.3480 \\
Rubber & Wood & 1.0000 & 0.0000 \\
SP 3K & SP 80 & 0.0370 & 0.9630 \\
SP 3K & Wood & 0.2907 & 0.7093 \\
SP 80 & Wood & 0.9444 & 0.0556 \\
\bottomrule
\end{tabular}
\end{minipage}
\hfill
\begin{minipage}[t]{0.48\textwidth}
\centering
\small
\captionof{table}{Cosmos2-2B Viscosity}
\begin{tabular}{llrr}
\toprule
\textbf{Group X} & \textbf{Group Y} & \textbf{P(X>Y)} & \textbf{P(Y>X)} \\
\midrule
Corn Syrup & Glycerine & 0.4298 & 0.5702 \\
Corn Syrup & Honey & 0.4139 & 0.5861 \\
Glycerine & Honey & 0.5344 & 0.4656 \\
\bottomrule
\end{tabular}
\end{minipage}

\vspace{2em}

\noindent
\begin{minipage}[t]{0.48\textwidth}
\centering
\small
\captionof{table}{Cosmos2.5-2B Friction}
\begin{tabular}{llrr}
\toprule
\textbf{Group X} & \textbf{Group Y} & \textbf{P(X>Y)} & \textbf{P(Y>X)} \\
\midrule
Plastic & Rubber & 0.0000 & 1.0000 \\
Plastic & SP 3K & 0.5710 & 0.4290 \\
Plastic & SP 80 & 0.0000 & 1.0000 \\
Plastic & Wood & 0.3630 & 0.6370 \\
Rubber & SP 3K & 1.0000 & 0.0000 \\
Rubber & SP 80 & 0.3801 & 0.6199 \\
Rubber & Wood & 0.9947 & 0.0053 \\
SP 3K & SP 80 & 0.0000 & 1.0000 \\
SP 3K & Wood & 0.2889 & 0.7111 \\
SP 80 & Wood & 1.0000 & 0.0000 \\
\bottomrule
\end{tabular}
\end{minipage}
\hfill
\begin{minipage}[t]{0.48\textwidth}
\centering
\small
\captionof{table}{Cosmos2.5-2B Viscosity}
\begin{tabular}{llrr}
\toprule
\textbf{Group X} & \textbf{Group Y} & \textbf{P(X>Y)} & \textbf{P(Y>X)} \\
\midrule
Corn Syrup & Glycerine & 0.5287 & 0.4713 \\
Corn Syrup & Honey & 0.4577 & 0.5423 \\
Glycerine & Honey & 0.4817 & 0.5183 \\
\bottomrule
\end{tabular}
\end{minipage}

\vspace{2em}

\noindent
\begin{minipage}[t]{0.48\textwidth}
\centering
\small
\captionof{table}{Hunyuan Friction}
\begin{tabular}{llrr}
\toprule
\textbf{Group X} & \textbf{Group Y} & \textbf{P(X>Y)} & \textbf{P(Y>X)} \\
\midrule
Plastic & Rubber & 0.0000 & 1.0000 \\
Plastic & SP 3K & 0.5726 & 0.4274 \\
Plastic & SP 80 & 0.0000 & 1.0000 \\
Plastic & Wood & 0.3761 & 0.6239 \\
Rubber & SP 3K & 1.0000 & 0.0000 \\
Rubber & SP 80 & 0.5585 & 0.4415 \\
Rubber & Wood & 1.0000 & 0.0000 \\
SP 3K & SP 80 & 0.0000 & 1.0000 \\
SP 3K & Wood & 0.3302 & 0.6698 \\
SP 80 & Wood & 1.0000 & 0.0000 \\
\bottomrule
\end{tabular}
\end{minipage}
\hfill
\begin{minipage}[t]{0.48\textwidth}
\centering
\small
\captionof{table}{Hunyuan Viscosity}
\begin{tabular}{llrr}
\toprule
\textbf{Group X} & \textbf{Group Y} & \textbf{P(X>Y)} & \textbf{P(Y>X)} \\
\midrule
Corn Syrup & Glycerine & 0.5531 & 0.4469 \\
Corn Syrup & Honey & 0.4897 & 0.5103 \\
Glycerine & Honey & 0.3870 & 0.6130 \\
\bottomrule
\end{tabular}
\end{minipage}

\vspace{2em}

\noindent
\begin{minipage}[t]{0.48\textwidth}
\centering
\small
\captionof{table}{Wan2.2 Friction}
\begin{tabular}{llrr}
\toprule
\textbf{Group X} & \textbf{Group Y} & \textbf{P(X>Y)} & \textbf{P(Y>X)} \\
\midrule
Plastic & Rubber & 0.0000 & 1.0000 \\
Plastic & SP 3K & 0.5598 & 0.4402 \\
Plastic & SP 80 & 0.0000 & 1.0000 \\
Plastic & Wood & 0.2704 & 0.7296 \\
SP 3K & SP 80 & 0.0000 & 1.0000 \\
SP 3K & Wood & 0.2077 & 0.7923 \\
SP 80 & Wood & 1.0000 & 0.0000 \\
\bottomrule
\end{tabular}
\end{minipage}
\hfill
\begin{minipage}[t]{0.48\textwidth}
\centering
\small
\captionof{table}{Wan2.2 Viscosity}
\begin{tabular}{llrr}
\toprule
\textbf{Group X} & \textbf{Group Y} & \textbf{P(X>Y)} & \textbf{P(Y>X)} \\
\midrule
Corn Syrup & Glycerine & 0.6053 & 0.3947 \\
Corn Syrup & Honey & 0.7204 & 0.2796 \\
Glycerine & Honey & 0.5844 & 0.4156 \\
\bottomrule
\end{tabular}
\end{minipage}

\clearpage

\end{document}